\documentclass[journal]{IEEEtran}
\usepackage{amsmath}
\usepackage{multirow}
\usepackage{graphicx}
\usepackage{float}
\usepackage{subfig}
\usepackage{algorithm}
\usepackage{algorithmic}
\usepackage{makecell}
\usepackage{bm}
\usepackage{hyperref} 
\usepackage{amssymb}
\ifCLASSINFOpdf
\else
\fi
\begin{document}
	
\captionsetup[Figure]{
	labelfont={bf},
	name={Fig.},
	labelsep=period
}
\captionsetup[table]{
	labelfont={bf},
	labelsep=newline,
	singlelinecheck=false,
}

%
\title{Differentiable SAR Renderer and SAR Target Reconstruction}
%
%
%

\author{Shilei~Fu,~\IEEEmembership{Student Member,~IEEE,}
        Feng~Xu,~\IEEEmembership{Senior Member,~IEEE}
\thanks{S. Fu and F. Xu are with the Key Laboratory of Information Science of Electromagnetic Waves (Ministry of Education), Fudan University, Shanghai 200433, China (e-mail: fengxu@fudan.edu.cn).}
}

%
%

\markboth{Journal of \LaTeX\ Class Files,~Vol.~14, No.~8, August~2015}%
{Shell \MakeLowercase{\textit{et al.}}: Bare Demo of IEEEtran.cls for IEEE Journals}
%



\maketitle

\begin{abstract}
Forward modeling of wave scattering and radar imaging mechanisms is the key to information extraction from synthetic aperture radar (SAR) images. Like inverse graphics in optical domain, an inherently-integrated forward-inverse approach would be promising for SAR advanced information retrieval and target reconstruction. This paper presents such an attempt to the inverse graphics for SAR imagery. A differentiable SAR renderer (DSR) is developed which reformulates the mapping and projection algorithm of SAR imaging mechanism in the differentiable form of probability maps. First-order gradients of the proposed DSR are then analytically derived which can be back-propagated from rendered image/silhouette to the target geometry and scattering attributes. A 3D inverse target reconstruction algorithm from SAR images is devised. Several simulation and reconstruction experiments are conducted, including targets with and without background, using both synthesized data or real measured inverse SAR (ISAR) data by ground radar. Results demonstrate the efficacy of the proposed DSR and its inverse approach.
\end{abstract}


%
\IEEEpeerreviewmaketitle

\section{Introduction}
%
%
%
%
\IEEEPARstart{S}{ynthetic} aperture radar (SAR) has become an important tool for Earth remote sensing \cite{IEEEhowto:Xu_GRSL}. It has the unique capability of high-resolution imaging regardless of weather and daylight. Due to the microwave frequency employed by and the phase-coherent nature of SAR imaging, a SAR image appears to be different from an optical one and is difficult to interpret. This has become a key bottleneck restricting wider applications of current spaceborne and airborne SAR systems \cite{IEEEhowto: MicrowaveVision}. The key to information retrieval and target reconstruction from SAR imagery is a deep understanding of the wave scattering physics and the unique SAR imaging mechanism, which has been the topic of many studies in the field of electromagnetic scattering modeling and SAR imaging simulation [e.g. \cite{IEEEhowto:MPA, IEEEhowto:BART}]. However, these forward models are often too complicated to be directly integrated into the inverse procedure of SAR information retrieval. Apparently, an inherently-integrated forward-inverse approach would be a promising approach to the advanced information retrieval (AIR) and target/scene reconstruction from SAR images \cite{IEEEhowto:Xu_GRSL}. In the computer vision regime, such method is referred to as 'inverse graphics' \cite{IEEEhowto:InverseGraphics}. This paper is an attempt to the inverse graphics for SAR imagery. We develop a differentiable renderer of SAR image which inherently integrates the forward SAR imaging process with the inverse target reconstruction algorithm.

Target geometry reconstruction from SAR image is the primary goal of AIR. For 3D reconstruction, conventional approaches resort to either interferometry or 3D imaging. For continuous earth surface such as mountainous area, single-baseline interferometric SAR (InSAR) can successfully derive the 3D elevation map by unwrapping the interferogram \cite{IEEEhowto:InSARReview}. For complex scenarios such as urban built-up areas, more advanced multi-baseline InSAR or tomographic SAR (TomoSAR) is capable of resolving the phase ambiguity due to abrupt elevation changes as well as extracting multiple overlapped scatters caused by the layover effect \cite{IEEEhowto:Zhu1}. Zhu et al. \cite{IEEEhowto:Zhu2} obtained a 3D reconstruction of Berlin, Germany from TomoSAR of over 450 TerrSAR-X images with a resolution of $1m$. TomoSAR often requires a large number of interferometric orbits which may take an extended period of time to acquire. In addition, it poses high requirements on the precision of phase error correction and thus hinders its wider applications. Ding et al. \cite{IEEEhowto: MicrowaveVision} proposed to take advantage of scattering and visual semantics as extracted from 2D SAR images and used them as strong regularizers for TomoSAR hoping to ease the requirements on the number of orbits.

In computer vision, data-driven deep learning approaches have achieved great successes, including the usage of deep neural networks to reconstruct 3D geometries from 2D images. Various representations for 3D geometries have been explored, such as depth images, voxels, point clouds, meshes, etc, among which, a mesh contains a vertex set and an edge set, and is suitable for graph-based convolutional neural networks (CNN) \cite{IEEEhowto:SurveyDGL}. Wang et al. \cite{IEEEhowto:Pixel2Mesh} proposed Pixel2Mesh to deform an ellipsoidal template to target mesh from a single 2D image. Wen et al. \cite{IEEEhowto:Pixel2MeshPlus} introduced a multi-view deformation network to the original Pixel2Mesh, and incorporated cross-view information in the process of mesh generation. Tang et al. \cite{IEEEhowto:Topologies} proposed a skeleton-bridged, stage-wise learning approach with good balance between topology preservation and low complexity.

Deep learning approaches have also been explored in SAR 3D reconstruction. For example, Peng et al. \cite{IEEEhowto:PointCloud} converted a single SAR image to optical perspective and recovered the 3D points based on a pre-trained 3D reconstruction network from the optical image. Wang et al. \cite{IEEEhowto:SingleTarget} trained a 3D super-resolution CNN to improve the resolution and the signal-to-noise ratio (SNR) of 3D reconstruction results when the number of observation orbits is insufficient. Chen et al. \cite{IEEEhowto:3dBuilding} proposed a coupled equivalent complex valued CNN for building facade detection in SAR images, and provided the detection results for the point cloud generation network to reconstruct 3D model of typical buildings.

These data-driven learning approaches often require a large number of training samples and is irrelevant of the actual imaging physics and mechanism, which are the subject of computer graphics. In computer graphics, an image can be rendered from a virtual object modeled by its geometry, appearance and attitude, and the illumination and camera configurations. As opposed to computer graphics that generates an image from the known parameters, the goal of computer vision is to extract the unknown parameters from the acquired image. Inverse graphics \cite{IEEEhowto:InverseGraphics} is a hybrid approach aiming to integrate the physics of computer graphics with the backward inference capability of machine learning. One way to realise inverse graphic is to develop a differentiable forward renderer of which the gradients can be calculated and used for error back-propagation. Note that error back-propagation is an effective inference algorithm that can be used to learn parameters of a forward model such as a neural network. However, conventional rendering involves a discrete operation named rasterization, which prevents error back-propagating from the obtained image to the unknown 3D object model. Recently, several differentiable renderers are being proposed. Open differentiable renderer (OpenDR) \cite{IEEEhowto:Opendr} approximates the gradients of 2D coordinates based on image derivatives by the local filtering operation. Neural 3D mesh renderer (NMR) \cite{IEEEhowto:Neural3D}, proposed by Kato et al., considers the pixel value to be a continuous function on the neighborhood coordinates of the facets. Soft rasterizer, proposed by Liu et al. \cite{IEEEhowto:SoftRas}, models the influence of triangular facets on the image by probability maps, and is also able to optimize the depth values of the traingles.

This paper develops a differentiable SAR renderer (DSR) which reformulates the mapping and projection algorithm (MPA) \cite{IEEEhowto:MPA} of SAR imaging mechanism while incorporating the soft rasterizer principles \cite{IEEEhowto:SoftRas}. An inverse target reconstruction algorithm based on DSR is then proposed and demonstrated with both simulated and real radar images. This paper implements a new framework for AIR that integrates, on the lower level, the SAR imaging mechanism with an inverse information retrieval algorithm. The main contributions of this paper are as follows:

\begin{enumerate}
	\item A novel differentiable SAR renderer: the differentiable MPA framework is formulated where the projection plane is established to account for the shadowing between facets by using rays, while the mapping plane is used to accumulate scattering contribution from facets onto imaging pixels. Both shadowing and scattering are accounted for in the form of differentiable probability maps.
	
	\item 3D inverse target reconstruction: the first-order gradients of the proposed DSR as back-propagated from rendered image/silhouette to the target geometric vertices and facet scattering attributes are derived, based on which a target reconstruction algorithm is devised.
	
	\item Extensive demonstration and evaluation: Several experiments are conducted, including targets with and without background. 3D target reconstruction from silhouettes is demonstrated using DSR-rendered, computationally simulated and radar measured SAR images. Reconstruction performance is quantitatively evaluated. Additional application of DSR is explored, such as predicting the attitude parameters from observed SAR images.
	
\end{enumerate}

The remainder of this paper is organized as follows. Section \uppercase\expandafter{\romannumeral2} presents the DSR forward model and derives the analytic forms of SAR images and silhouettes as a function of target geometry and radar configuration. Then in Section \uppercase\expandafter{\romannumeral3}, a new inverse reconstruction framework is introduced where the loss function is established and the gradients for back-propagation are derived. Rendering and reconstructing experiments on various models are conducted in Section \uppercase\expandafter{\romannumeral4} to verify the effectiveness of DSR. Section \uppercase\expandafter{\romannumeral5} discusses time consumption of forward rendering and backward reconstruction. Finally, Section \uppercase\expandafter{\romannumeral6} concludes the paper.

\section{Forward Rendering}
\begin{figure}[!t]
	\centering
	\includegraphics[height=2.2in]{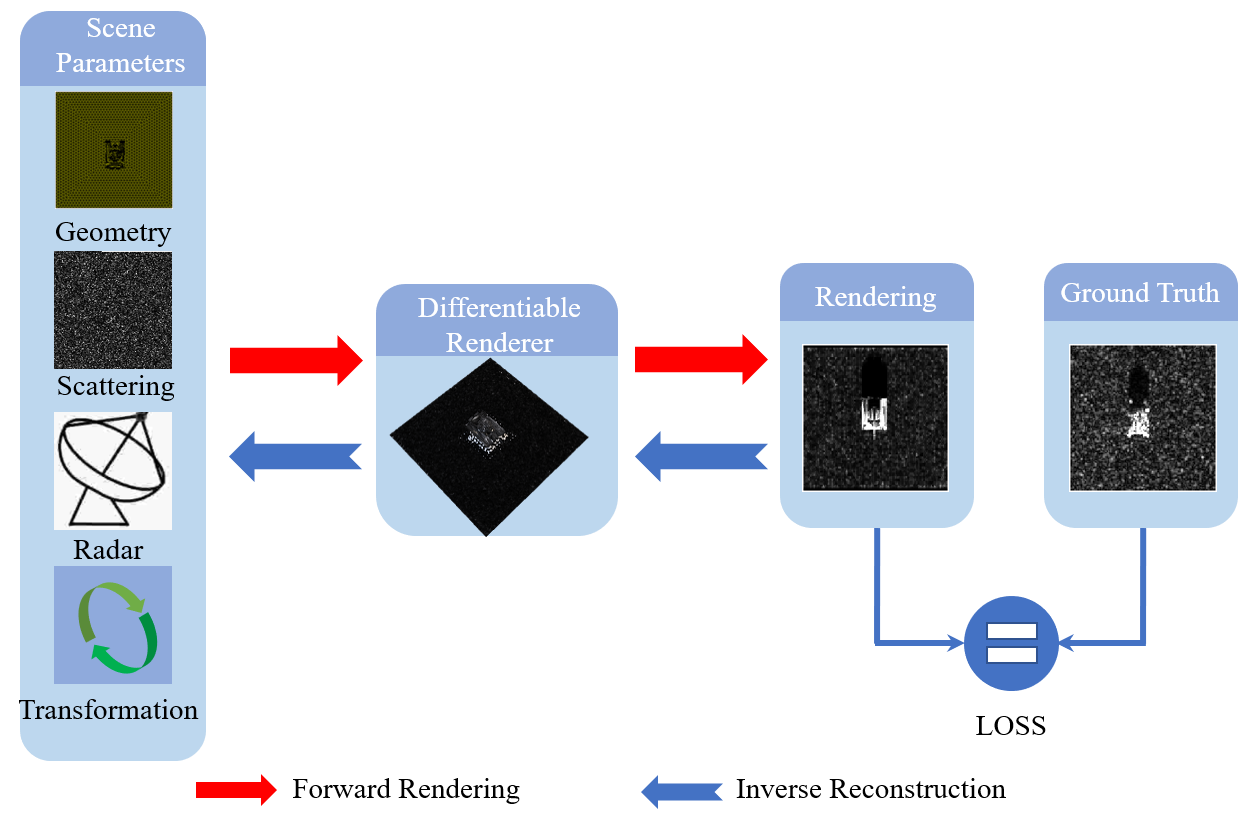}	
	\caption{The overall framework of DSR.}
	\label{figure29}
\end{figure}

A differentiable SAR renderer, abbreviated to DSR, is proposed in this paper. A SAR image can be seen as a function of target geometry, facet scattering attributes, radar imaging configuration as shown in \autoref{figure29}. The facet attribute of a mesh should be the scattering matrix which is essentially governed by the facet's material property and surface roughness. The radar replaces the camera as the sensor, and SAR has a very different imaging projection scheme from that of camera. SAR image is azimuth vs. slant-range parallel projection where the resolution does not vary with distance.

\begin{figure}[!t]
	\centering
	\includegraphics[height=1.6in]{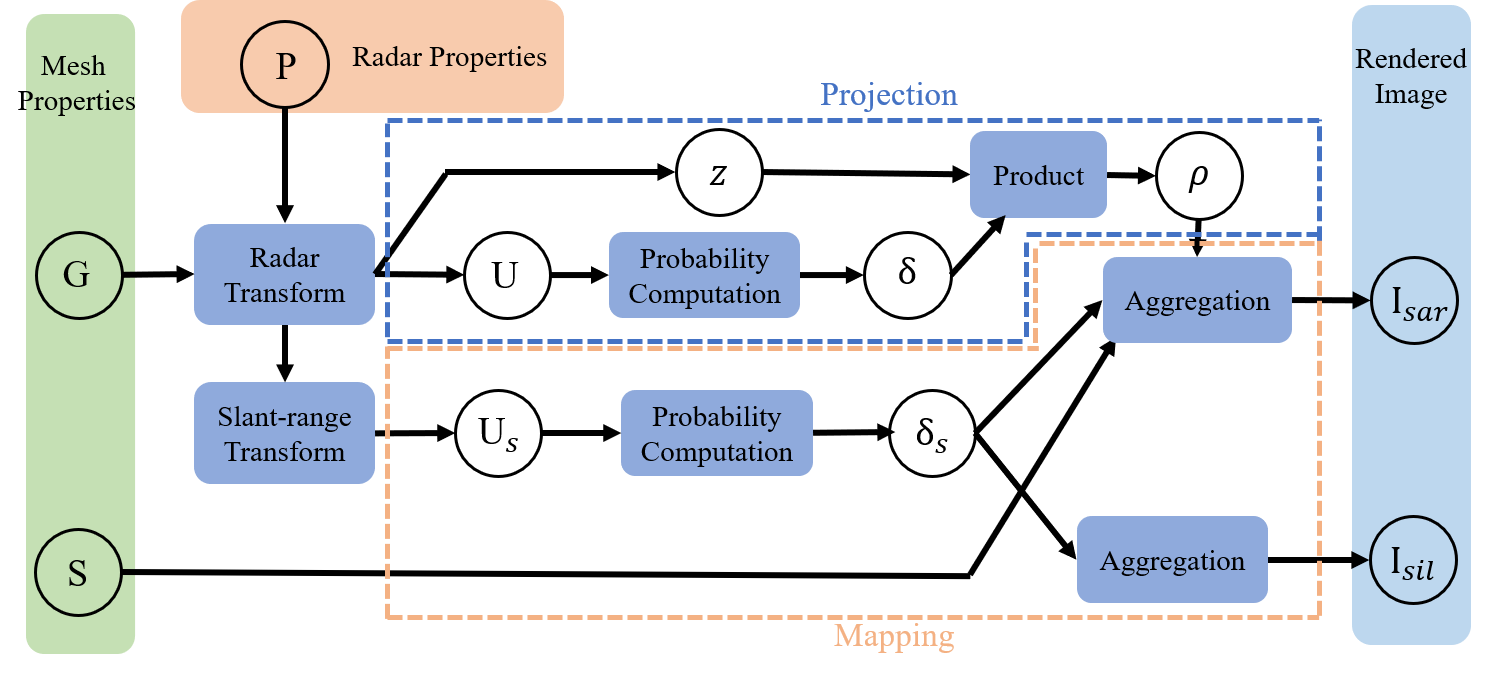}	
	\caption{Rendering framework.}
	\label{figure1}
\end{figure}

As shown in \autoref{figure1}, extrinsic variables (radar position \textbf{P}) define the radar sensing configuration, and intrinsic properties (mesh geometry \textbf{G} and per-facet scattering value \textbf{S}) describe the model-specific attributes. Following the rendering pipeline, image-space coordinate \textbf{U} and depth \textbf{Z} are obtained by transforming the input geometry \textbf{G} according to radar \textbf{P}. Due to slant-range imaging mechanism, subsequent slant-range transformation is adopted, and gets \bm{${\rm U_s}$}. Different from traditional rasterization formulated as discrete binary masks (shown later in \autoref{figure2}(a)), we use probability map \bm{${\rm \delta}$} which models the probability of each pixel staying inside each triangle facet \cite{IEEEhowto:SoftRas}. Model the shadowing relationship between facets as \bm{${\rm \rho}$}. \bm{${\rm \mathbf{I}_{sar}}$} is an aggregation function that fuses occlusion probability \bm{${\rm \rho}$}, per-facet scattering maps \bm{${\rm S}$} and probability map \bm{${\rm \delta_s}$}. \bm{${\rm \mathbf{I}_{sil}}$} is related to \bm{${\rm \delta_s}$}, and is actually an aggregation function for the silhouette.

Meshes are the most commonly used format to depict high quality 3D shapes with less memory and computational cost. A mesh contains a vertex set $\left\{ \mathbf{v}^{o}_{1}, \mathbf{v}^{o}_{2}, \cdots, \mathbf{v}^{o}_{N_v} \right\}$ and a facet set $\left\{ \mathbf{f}_{1}, \mathbf{f}_{2}, \cdots, \mathbf{f}_{N_f} \right\}$, where the object has $N_v$ vertices and $N_f$ facets. $\mathbf{v}^{o}_{i} \in \mathbb{R} ^ {3 \times 1}$ represents the spatial position of the $i$-th vertex and $\mathbf{f}_j \in \mathbb{N}_{+} ^ {3 \times 1}$ denotes the indices of the three vertices that belongs to the $j$-th triangular facet. Vertices $\left\{ \mathbf{v}^{o}_{i} \right\}$ can be transformed to vertices $\left\{ \mathbf{v}_{i} \right\}$ in the radar coordinate system. Every triangular facet $\mathbf{f}_{j}$ has one attribute, texture (scattering value) $S_j$. Note that single polarized SAR images are rendered in this paper, so $S_j$ is a scalar here. Its geometry (vertex coordinates) $\mathbf{M}_j$ after radar transformation could also expressed as

\begin{equation}
	\label{equ_6}
	\mathbf{M}_{j}=\left[\begin{array}{lll}
	x_{j, 1} & y_{j, 1} & z_{j, 1} \\
	x_{j, 2} & y_{j, 2} & z_{j, 2} \\
	x_{j, 3} & y_{j, 3} & z_{j, 3}
	\end{array}\right]
\end{equation}
where $(x_{j,n}, y_{j,n}, z_{j,n})$ is $x, y, z$ coordinates of the $n$-th vertex of $\mathbf{f}_j$. 

Relevant variables about the mesh and the coordinate system are summarized in \autoref{table9}.
\begin{table}[!t]
	\caption{Variable notations}
	\label{table9} 
	
	\renewcommand{\arraystretch}{2}
	\setlength{\tabcolsep}{1.2mm}
	\centering
	
	\begin{tabular}{p{1.0cm} p{1.0cm} p{6.5cm}}
		\hline \hline
		\textbf{Notation} & \textbf{Domain} & \textbf{Description} \\
		\hline
		$\mathbf{p}^{(i,l)}$ & $\mathbb{R}^{3 \times 1}$ & Coordinate vector of the $i$-row and $l$-column projection cell. \\
		\hline
		$\mathbf{m}^{(k,l)}$ & $\mathbb{R}^{3 \times 1}$ & Coordinate vector of the $k$-row and $l$-column mapping cell. \\
		\hline
		$\mathbf{f}_j$ & $\mathbb{N}_{+}^{3 \times 1}$ & Index vector of the three vertices in the $j$-th triangular facet.\\
		\hline
		$S_j$ & $\mathbb{R}$ & Scalar of the scattering value in the $j$-th triangular facet. (\bm{${\rm S}$} in \autoref{figure1}) \\
		\hline
		$\delta_{j}^{(i,l)}$ & $\mathbb{R} $ & Scalar of occupation probability of $\mathbf{f}_{j}$ on $\mathbf{p}^{(i,l)}$. (\bm{${\rm \delta}$} in \autoref{figure1}) \\
		\hline
		$\delta_{j}^{(k,l)}$ & $\mathbb{R} $ & Scalar of occupation probability of $\mathbf{f}_{j}$ on $\mathbf{m}^{(k,l)}$. (\bm{${\rm \delta_s}$} in \autoref{figure1}) \\
		\hline
		$z_{j}^{(i,l)}$ & $\mathbb{R} $ & Scalar of the normalized depth of $Z_{j}^{(i,l)}$. (\bm{${\rm z}$} in \autoref{figure1})\\
		\hline
		$\rho_{j}^{(i,l)}$ & $\mathbb{R} $ & Energy scalar of $\mathbf{f}_{j}$ allocated by $\mathbf{p}^{(i,l)}$. (\bm{${\rm \rho}$} in \autoref{figure1})\\
		\hline
		$\omega_{j}^{(k,l)}$ & $\mathbb{R} $ & Energy scalar of $\mathbf{m}^{(k,l)}$ reflected by $\mathbf{f}_{j}$. \\
		\hline
		$I_{sar}^{(k,l)}$ & $\mathbb{R}$ & Scattering intensity at the cell $\mathbf{m}^{(k,l)}$. (\bm{${\rm \mathbf{I}_{sar}}$} in \autoref{figure1}) \\
		\hline
		$I_{sil}^{(k,l)}$ & $\mathbb{R}$ & Silhouette value at the cell $\mathbf{m}^{(k,l)}$. (\bm{${\rm \mathbf{I}_{sil}}$} in \autoref{figure1}) \\
		\hline \hline
	\end{tabular}
\end{table}

\subsection{Coordinate Systems}
\begin{figure}[!t]
	\centering
	\includegraphics[height=2.0in]{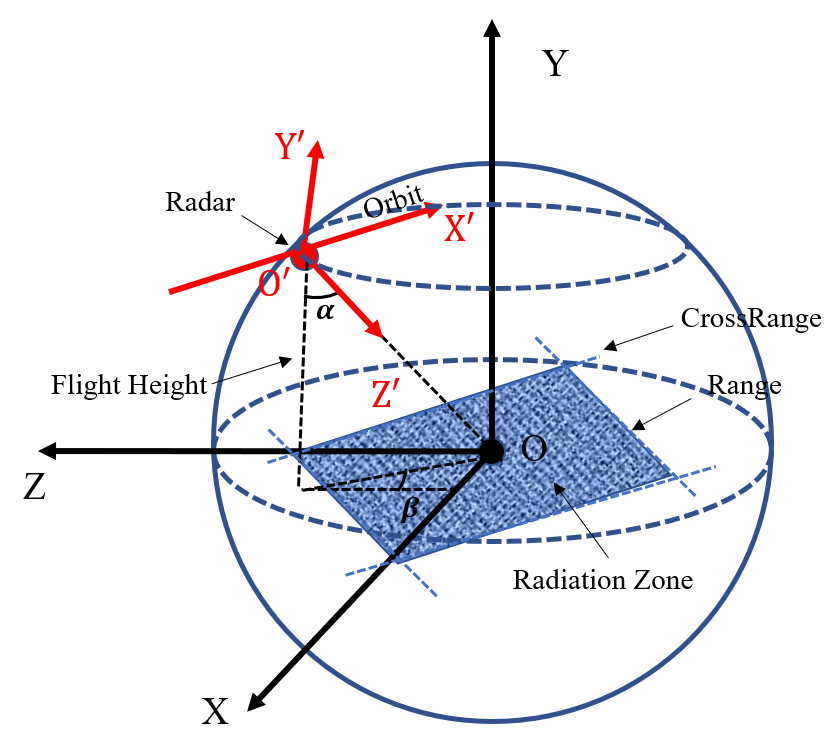}	
	\caption{The definition of related coordinate systems.}
	\label{figure4}
\end{figure}

In the world coordinate system $O$-$XYZ$, the imaging target is placed at the origin $O$. It is known that the nominal position of radar is at point $O'$, whose coordinate is $(x_r,y_r,z_r)$. The radar’s movement direction is $O'X'$ and slant range direction (looking direction) is $O'Z'$. Hence, the 3rd axis $O'Y'$ can be solely determined, so as the radar coordinate system $O'$-$X'Y'Z'$. The plane $O'Z'X'$ is defined as the mapping plane, i.e. the imaging plane, and the plane $O'X'Y'$ is defined as the projection plane, i.e. the shadowing plane. The projection plane is used to describe the shadowing relationship between facets, and the mapping plane is used to accumulate the scattering of facets onto the SAR image. In \autoref{figure4}, $\alpha$, $\beta$ represent the incident angle and azimuth angle respectively.

\textbf{Radar Transform} The world coordinate system can be converted to the radar coordinate system through the rotation matrix $\mathbf{R}$, whose column is the unit directional vector of each axis of the system $O'$-$X'Y'Z'$, i.e.

\begin{equation}
	\label{equ_1}
	\mathbf{R}=\left[\begin{array}{ccc}
	-\cos \beta & -\cos \alpha \sin \beta & -\sin \alpha \sin \beta \\
	0 & \sin \alpha & -\cos \alpha \\
	\sin \beta & -\cos \alpha \cos \beta & -\sin \alpha \cos \beta
	\end{array}\right]
\end{equation}

Thus, any vertex of the mesh can be transformed to the radar coordinate system as
\begin{equation}
	\label{equ_2}
	\mathbf{v}_r = \mathbf{R} ^ {\mathbf{T}} {\left( \mathbf{v} - \mathbf{p}_r \right)}  
\end{equation}
where $\mathbf{p}_r$ denotes the radar's antenna phase center position. $\mathbf{v}$ and $\mathbf{v}_r$ denote the coordinates of the same vertex in the world and the radar coordinate systems, respectively.

\begin{figure}[!t]
	\centering
	\includegraphics[height=2.0in]{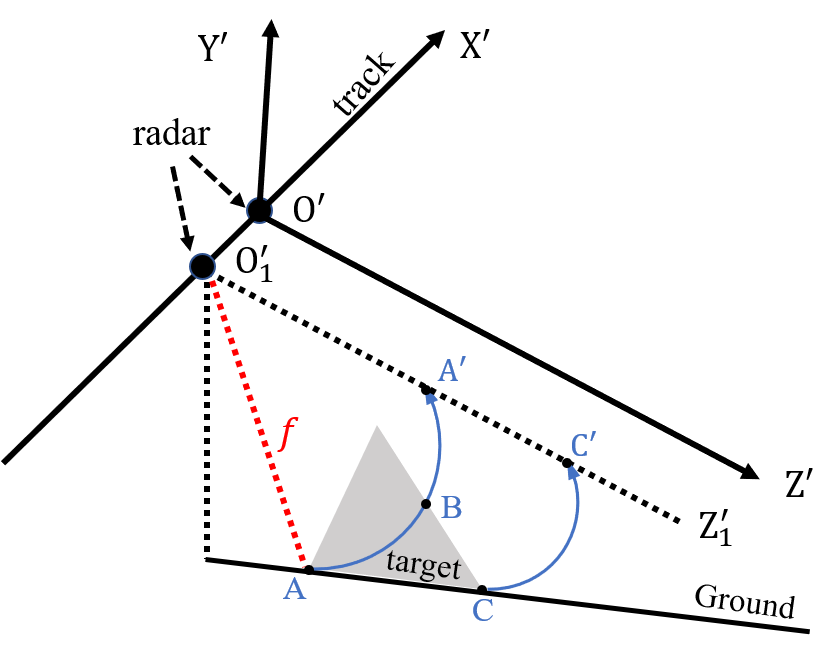}	
	\caption{Slant-range transform.}
	\label{figure5}
\end{figure}

\textbf{Slant-range Transform} As shown in \autoref{figure5}, when performing SAR imaging, slant-range mapping is used. $O'X'$ is the orbital direction of the radar, that is, the azimuth direction. $O'Z'$ is the direction from the radar to the center of the target, also known as slant-range direction. When the radar is at point $O'_1$, all scatterers on the vertical plane $X'=O'_1$ will be mapped to the line $O'_1 Z'_1$. For example, points $A$ and $B$, with the same distance from $O'_1$, will be mapped to the same point $A'$. This is the layover effect. Specifically, for any point $A\left( x,y,z \right)$, it first

(1) maps the point on the $O'Y'Z'$ plane to the  range $O'Z'$ by

\begin{equation}
	\label{equ_3}
	z' = \sqrt{y ^ 2 + z ^ 2}
\end{equation}

(2) subtracts the reference range as

\begin{equation}
	\label{equ_4}
	\hat{z} = z' - f
\end{equation}
where, $f$ is the range from the radar to the scene center. So the projection point $A'$ will be $(x, 0, \hat{z})$.

\begin{figure}[!t]
	\centering
	\includegraphics[height=2.0in]{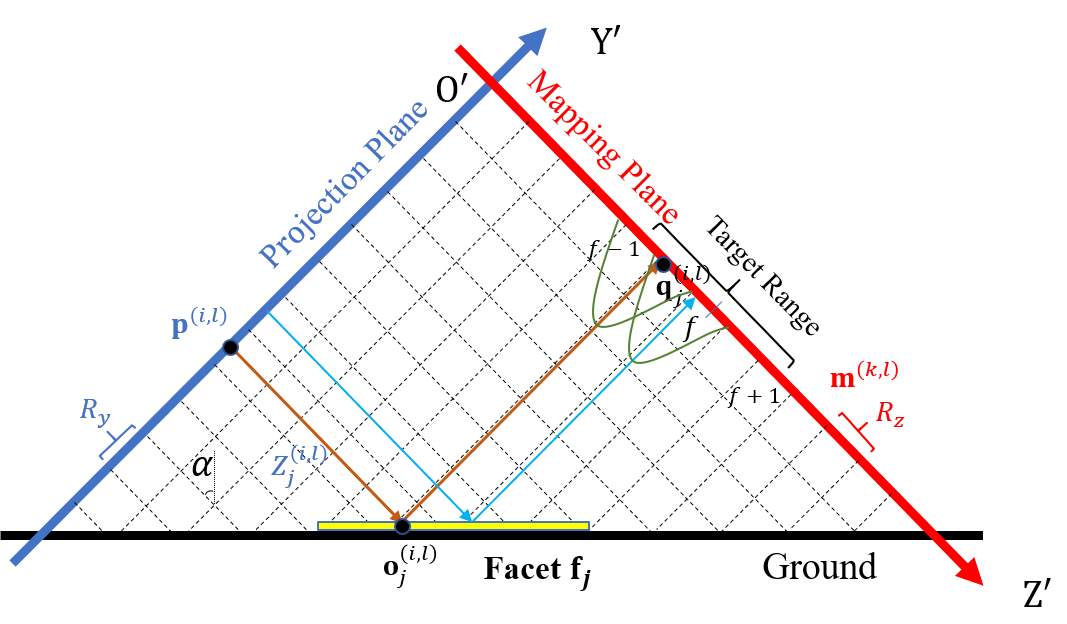}	
	\caption{The mapping and projection algorithm.}
	\label{figure6}
\end{figure}

In the MPA (see \autoref{figure6}), the imaging plane is discretized as many pixels, called mapping cells, and the $(k,l)$-th cell is denoted as $\mathbf{m}^{(k,l)}$, whose size in $Z'$ dimension equals to the range resolution $R_z$. Likewise, the projection plane can be also uniformly discretized as $N_y \times N_x$ projection cells and the $(i,l)$-th cell is denoted as $\mathbf{p}^{(i,l)}$, whose size in $Y'$ direction is denoted as $R_y$.

The purpose of projection is to calculate the mutual shadowing effects among facets. It can be seen as casting a grid of rays from the source screen of $O'X'Y'$ and the rays will hit facets in turn along the $O'Z'$ direction. Hence, it can determine one facet occludes the other. $R_y$ is projection cell size, corresponding to the granularity of ray grid. It determines the accuracy of the facet occlusion. We can relate the projection cell size with the imaging resolution by projecting $R_y$ onto the ground plane as $ R_y / \cos{\alpha}$ and assure it equal with the slant range resolution $ R_z / \sin{\alpha}$, which yields $R_y=R_z \cot{\alpha }$. Assuming that the number of resolution cells along $O'Z'$ direction is $N_z$, according to the relationship between $R_y$ and $R_z$, the number of projection cells along $O'Y'$ can be obtained as $N_y = \left\lceil N_z \tan{\alpha} \right\rceil$. Note that as the incident angle increases, $N_y$ will increase rapidly.

\subsection{Differentiability}
Forward rendering for camera or SAR is in fact the mapping from 3D scene to 2D image. If the process can be reversed, the coordinates and textures of the target may be inferred. However, traditional graphic renderer includes two discrete steps: rasterization and depth buffering, making the rendering process non-differentiable. We can convert these discrete operations to continuous ones in the form of probability maps \cite{IEEEhowto:SoftRas}.

\begin{figure}[!t]
	\centering
	\subfloat[Rasterization of a triangle]{
		\includegraphics[height=1.0in]{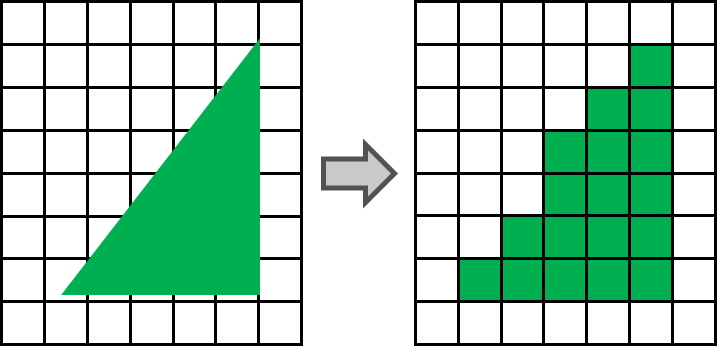}
	}
	\subfloat[Definition of pixel-to-triangle distance.]{
		\includegraphics[height=1.0in]{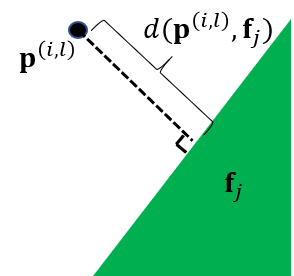}
	}
	\quad
	\subfloat[Probability map with $\sigma=0.1$.]{
		\includegraphics[width=1.6in]{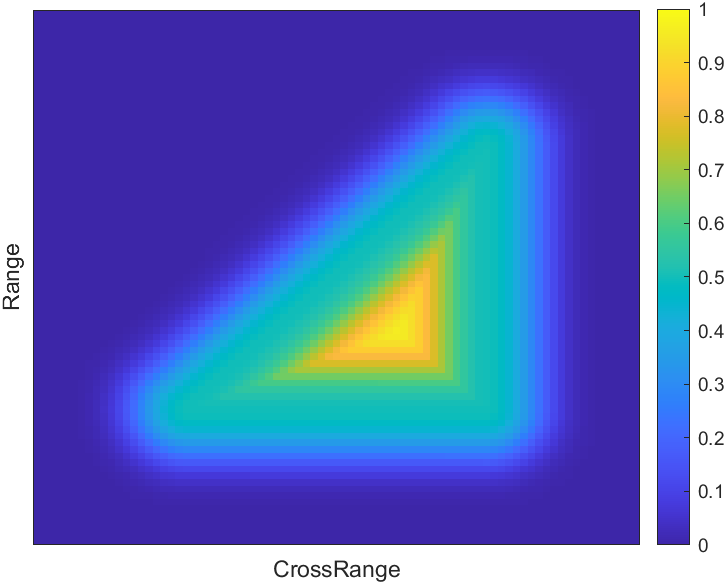}
	}
	\subfloat[Probability map with $\sigma=1e-5$.]{
		\includegraphics[width=1.6in]{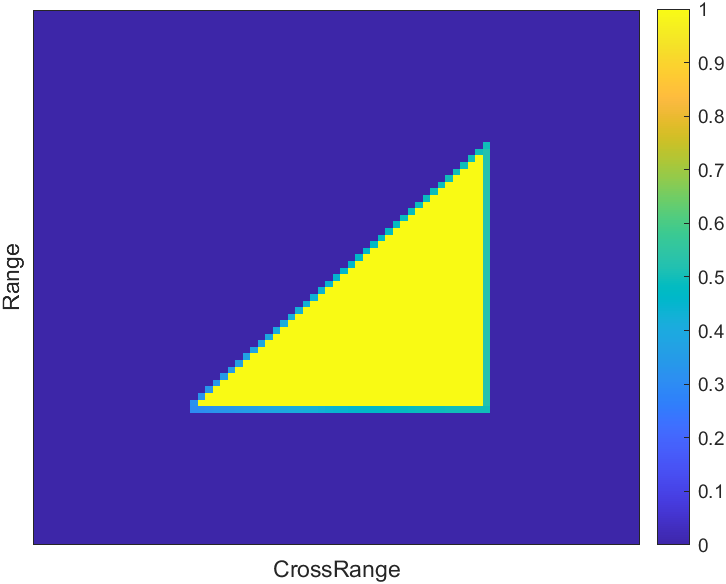}
	}
	\caption{Comparison between rasterization and probability calculation.}
	\label{figure2}
\end{figure}

Rasterization converts vertex coordinates to pixel indices. Only binary relationships exist between image pixels and facets: a pixel inside the facet or outside. We adopt probability map computation instead of traditional rasterization \cite{IEEEhowto:SoftRas}. We model the influence of triangle $\mathbf{f}_{j}$ on image plane by probability map ${\bm \delta}_j$. We define ${\bm \delta}_j$ at pixel $\mathbf{p}^{(i,l)}$ as follows:

\begin{equation}
	\label{equ_7}
	\delta_{j}^{(i,l)}=\frac{1}{1+\exp \left(-s^{(i,l)}_{j} \cdot d\left(\mathbf{p}^{(i,l)}, \mathbf{f}_{j}\right)^{2} / \sigma\right)}
\end{equation}
where $s^{(i,l)}_{j}$ is a sign indicator whether the cell is inside or outside $\mathbf{f}_{j}$, that is $s^{(i,l)}_{j}=\left\{ +1,\text{if} \, \mathbf{p}^{(i,l)}\in \mathbf{f}_{j}; -1, \text{otherwise} \right\}$. $d\left( \mathbf{p}^{(i,l)},\mathbf{f}_{j} \right)$ is the closest Euclidean distance from $\mathbf{p}^{(i,l)}$ to $\mathbf{f}_{j}$'s edges in \autoref{figure2}(b). 

$\sigma$ is a scalar that controls the sharpness of probability distribution. When $\sigma=0.1$ shown in \autoref{figure2}(c), the probability map gradually changes from 0 to 1 as it moves from outside to inside the facet. As $\sigma$ decreases, the transition across the edges of facet become sharper, and the boundary is clearer. When $\sigma \to 0$, the probability map converges to the exact shape of the facet boundary (see \autoref{figure2}(d)).

Let the vertex set of facet $\mathbf{f}_{j}$ be $\left\{ \mathbf{v}_{1}, \mathbf{v}_{2}, \mathbf{v}_{3} \right\}$. Denote the closet point on the edge $\left( \mathbf{v}_{k+1} - \mathbf{v}_{k} \right)$ to $\mathbf{p}^{(i,l)}$ as $\mathbf{p}_{k}$. Here $\left( \mathbf{v}_{k+1} - \mathbf{v}_{k} \right)$ denotes a vector with $\mathbf{v}_{k+1}$ as the starting point and $\mathbf{v}_{k}$ as the end point. Then the distance vector $\left(  \mathbf{p}_{k} - \mathbf{p}^{(i,l)} \right)$ satisfies

\begin{equation}
	\label{equ_10}
	\begin{aligned}
	& \mathbf{p}_{k} - \mathbf{p}^{(i,l)} = \mathbf{v}_{k} - \mathbf{p}^{(i,l)}  + \left( \mathbf{v}_{k+1} - \mathbf{v}_{k} \right)  \cdot \\ 
	& \qquad  \max{\left\{ \min{\left(\frac{ \left( \mathbf{v}_{k} - \mathbf{p}^{(i,l)} \right)  \cdot \left(\mathbf{v}_{k+1} - \mathbf{v}_{k} \right)}{\left(\mathbf{v}_{k+1} - \mathbf{v}_{k} \right) \cdot \left(\mathbf{v}_{k+1} - \mathbf{v}_{k} \right)}, 1\right)}, 0 \right\}}
	\end{aligned}
\end{equation}

The closet point $\mathbf{t}$ from the cell to the facet is the point with the smallest magnitude, i.e.

\begin{equation}
	\label{equ_10}
	\mathbf{t}=\arg \min_{\mathbf{p}_{k}} \left\{ \left| \mathbf{p}_{k} - \mathbf{p}^{(i,l)} \right| \right\}
\end{equation}

$d\left( \mathbf{p}^{(i,l)},\mathbf{f}_{j} \right)$ is defined as the distance between $\mathbf{t}$ and $\mathbf{p}^{(i,l)}$.
\begin{equation}
	\label{equ_8}
	d\left(\mathbf{p}^{(i,l)},\mathbf{f}_{j} \right)
	=\left\| \mathbf{t} - \mathbf{p}^{(i,l)} \right\|_{2}
	=\left\|\mathbf{U}_{j}\left(\mathbf{t}_{j}^{(i,l)}-\mathbf{b}_{j}^{(i,l)}\right)\right\|_{2} \\
\end{equation}
where $\mathbf{b}_{j}^{(i,l)}=\left\{ b_{j,1}^{(i,l)}, b_{j,2}^{(i,l)}, b_{j,3}^{(i,l)} \right\}$ is the barycentric coordinate of the cell $\mathbf{p}^{(i,l)}$ in the system defined by $\mathbf{f}_{j}$. And  $\mathbf{t}_{j}^{(i,l)}=\left\{ t_{j,1}^{(i,l)}, t_{j,2}^{(i,l)}, t_{j,3}^{(i,l)} \right\}$ is that of point $\mathbf{t}$. The barycentric coordinate system is a reference system constructed by the three vertices of the triangular facet. For example, the barycentric coordinate of $\mathbf{p}^{(i,l)}$ satisfies \cite{IEEEhowto:SoftRas}

\begin{equation}
	\label{equ_9}
	\mathbf{b}_{j}^{(i,l)}=\mathbf{U}_{j}^{-1} \hat{\mathbf{p}}^{(i,l)}=\left[\begin{array}{ccc}
	x_{j,1} & x_{j,2} & x_{j,3} \\
	y_{j,1} & y_{j,2} & y_{j,3} \\
	1 & 1 & 1
	\end{array}\right]^{-1}\left[\begin{array}{c}
	x^{(i,l)} \\
	y^{(i,l)} \\
	1 \\
	\end{array}\right]
\end{equation}
where $\mathbf{U}_{j}$ is introduced as an intermediate variable indicating 2D projected vertex coordinate positions of $\mathbf{f}_{j}$. $\mathbf{p}^{(i,l)}$'s coordinate is $(x^{(i,l)}, y^{(i,l)}, z^{(i,l)})$. Set $z^{(i,l)}=1$ to project $\mathbf{p}^{(i,l)}$ onto the same plane as $\mathbf{U}_{j}$ and get $\hat{\mathbf{p}}^{(i,l)}$.

\begin{figure}[!t]
	\centering
	\includegraphics[height=1.8in]{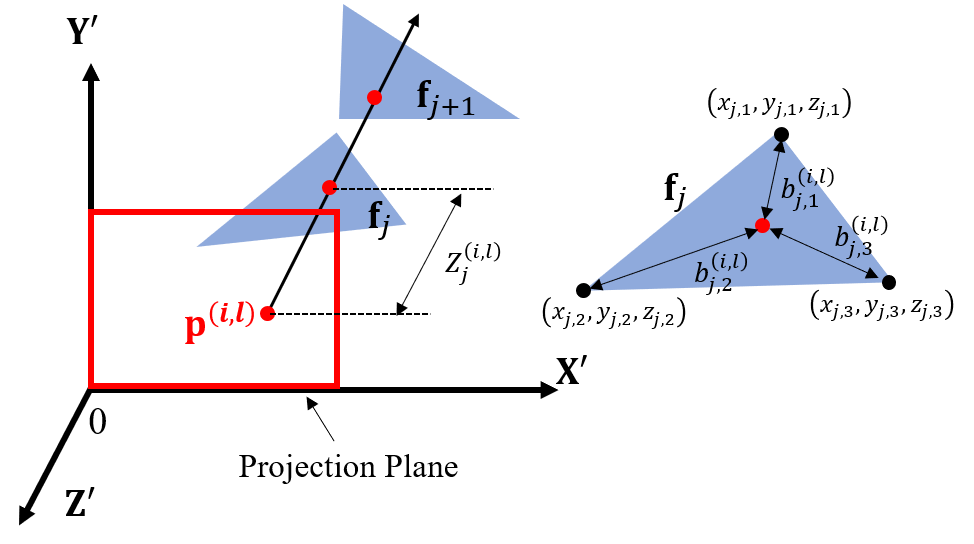}	
	\caption{Schematic diagram of depth buffering.}
	\label{figure3}
\end{figure}

During depth buffering shown in \autoref{figure3}, both facets $\mathbf{f}_{j}$ and $\mathbf{f}_{j+1}$ are projected onto point $\mathbf{p}^{(i,l)}$ on the projection plane. Since facet $\mathbf{f}_{j}$ is closer, $\mathbf{f}_{j+1}$ is shadowed by $\mathbf{f}_{j}$ and could not contribute to the image. We use $z_j^{(i,l)}$, the normalized depth of $\mathbf{p}^{(i,l)}$ on $\mathbf{f}_{j}$, as a weight to realize the shadowing effect, so that all triangles have probabilistic contributions to each cell. The closer $\mathbf{f}_{j}$ is to $\mathbf{p}^{(i,l)}$, the larger $z_j^{(i,l)}$ will be.

\begin{equation}
	\label{equ_10}
	z_{j}^{(i,l)}=\frac{Z_{f}-Z_{j}^{(i,l)}}{Z_{f}-Z_{n}}
\end{equation}
where $Z_f$ and $Z_n$ denote the far and near cut-off distances of radar field. $Z_j^{(i,l)}$ denotes the actual distance between $\mathbf{p}^{(i,l)}$ and $\mathbf{f}_{j}$ in $O'Z'$ direction, satisfying

\begin{equation}
	\label{equ_11}
	\frac{1}{Z_{j}^{(i,l)}}={\sum_{n=1}^{3} \frac{b_{j, n}^{(i,l)}}{z_{j, n}}}
\end{equation}

\subsection{Projection}
A projection cell $\mathbf{p}^{(i,l)}$, viewed as a ray generated by $(i,l)$-th cell on the projection plane, hits the facet $\mathbf{f}_{j}$ at point $\mathbf{o}_j^{(i,l)}$ (shown in \autoref{figure6}). The ray intensity received by $\mathbf{f}_{j}$ relates to the intersection probability $\delta_{j}^{(i,l)}$ on the plane and the relative distance $Z_{j}^{(i,l)}$ in the depth direction. The intensity of $\mathbf{f}_j$ allocated by $\mathbf{p}^{(i,l)}$ is then denoted as

\begin{equation}
	\label{equ_12}
	\rho_{j}^{(i,l)}=\frac{\delta_{j}^{(i,l)} \exp \left(z_{j}^{(i,l)} / \gamma\right)}{\sum_{k} \delta_{k}^{(i,l)} \exp \left(z_{k}^{(i,l)} / \gamma\right)}
\end{equation}
where $\gamma$ is a scalar, controlling the degree of occlusion. As $\gamma \to 0$, the far point will be completely blocked by the near point, which is equivalent to depth buffering. The purpose of normalization is to make the energy of each ray equal to 1.

The energy of $\mathbf{o}_j^{(i,l)}$ reflects to the point $\mathbf{q}_j^{(i,l)}$. The reflection path is parallel to $O'Y'$ axis. Then for the cell $\mathbf{m}^{(k,l)}$ with the same azimuth angle on the mapping angle ($l$-th column), the energy obtained from the point $\mathbf{q}_j^{(i,l)}$ can be formulated as

\begin{equation}
	\label{equ_13}
	\omega_{j}^{(k,l) \mid (i,l)}=\rho_{j}^{(i,l)} \cdot f\left(d_{z}\left(\mathbf{q}_{j}^{(i,l)}, \mathbf{m}^{(k,l)}\right)\right)
\end{equation}
where $d_z\left(\mathbf{q}_j^{(i,l)},\mathbf{m}^{(k,l)} \right)$ is the distance between $\mathbf{q}_j^{(i,l)}$ and $\mathbf{m}^{(k,l)}$ along $O'Z'$ direction, following

\begin{equation}
	\label{equ_14}
	d_z\left(\mathbf{q}_j^{(i,l)},\mathbf{m}^{(k,l)} \right) = Z^{(i,l)}_{j} + z^{(k,l)}  - f
\end{equation}
where $z^{(k,l)}$ is $z$- direction coordinate of $\mathbf{m}^{(k,l)}$.

$f(\cdot)$ denotes the energy distribution function, which distributes energy according to the distance from the central point. Here, a Gaussian function is used.

\begin{equation}
	\label{equ_15}
	f(d) =\frac{\exp \left(-d ^ {2} / 2 \sigma_g ^ {2}\right)}{\sqrt{2 \pi} \sigma_g}
\end{equation}
where $d$ denotes the distance between two points. $\sigma_g$ is the standard deviation of $f(\cdot)$. As $\sigma_g$ increases, the probability curve flattens. Points far from the center can be allocated more energy, and the total amount of energy allocated to the limited sampling cells becomes less due to the energy lost outside the target region.

$\mathbf{m}^{(k,l)}$ finally gets accumulated energy from all rays on the projection plane reflected by facet $\mathbf{f}_{j}$ as

\begin{equation}
	\label{equ_16}
	\omega_{j}^{(k,l)}=\sum_{i} \omega_{j}^{(k,l) \mid (i,l)}
\end{equation}

\subsection{Mapping}
The mapping plane is $O'Z'X'$. The coordinates in the radar coordinate system need to be mapped to the mapping plane through slant-range transform. When texturing the cell $\mathbf{m}^{(k,l)}$, the contribution of all facets should be considered. The contribution of facet $\mathbf{f}_{j}$ is related to the ray intensity $\omega_j^{(k,l)}$ and the occupation $\delta _j^{(k,l)}$ on the plane, that is

\begin{equation}
	\label{equ_18}
	I_{sar, j}^{(k,l)} \propto \delta_{j}^{(k,l)} \cdot \omega_{j}^{(k,l)}
\end{equation}
where $\delta_{j}^{(k,l)}$ is calculated based on \autoref{equ_7}, but expanded on the mapping plane. It denotes the influence of facet $\mathbf{f}_{j}$ after slant-range transformation on the cell $\mathbf{m}^{(k,l)}$.

As the scattering value of $\mathbf{f}_{j}$ is $S_j$, the scattering intensity $I_{sar}^{(k,l)}$ at the cell $\mathbf{m}^{(k,l)}$ can be obtained, i.e.

\begin{equation}
	\label{equ_19}
	I_{sar}^{(k,l)}=\sum_{j=1}^{N_f} \delta_{j}^{(k,l)} \cdot \omega_{j}^{(k,l)} \cdot S_{j}
\end{equation}
where $N_f$ is the number of facets.

We also explore the aggregate function for silhouette as

\begin{equation}
	\label{equ_20}
	I_{sil}^{(k,l)}=1-\prod_{j=1}^{N_f}\left(1-\delta_{j}^{(k,l)}\right)
\end{equation}

Silhouette $\mathbf{I}_{sil}$ is the probability of having at least one facet contributing to each mapping cell. It is independent of the textures and depths of facets. It is proposed for more easily calculating the gradients of geometric coordinates without considering textures and depth maps.

\subsection{Rendering Algorithm}
If the energy $\rho_{j}^{(i,l)}$ of facet $\mathbf{f}_{j}$ from the $(i,l)$-th ray generated by the projection plane is recorded in the graphic memory, it takes about $O(N_y \cdot N_x \cdot N_f)$ memory space. When $N_y$ increases sharply with the increase of the incident angle, the memory required also increases rapidly, resulting in insufficient graphic memory. This can be resolved by trading computation time for memory. Let $I_{sar}^{(k,l)}$ be transformed to the following \autoref{equ_21}, and repeatedly calculate $\rho_j^{(i,l)}$ so that the required memory can be reduced to $O(N_z \cdot N_x \cdot N_f)$.

\begin{equation}
	\label{equ_21}
	I_{sar}^{(k,l)}=\sum_{i} \sum_{j} \delta_{j}^{(k,l)} \cdot S_{j} \cdot \omega_{j}^{(k,l) \mid (i,l)}
\end{equation}

As shown in Algorithm 1, the goal is to render two images $\left\{ I_{sar}^{(k,l)} \right\}$ and $\left\{ I_{sil}^{(k,l)} \right\}$. Among them, $\left\{ I_{sil}^{(k,l)} \right\}$ only relates to $\delta_j^{(k,l)}$, and needs to traverse all facets; meanwhile, $\left\{ I_{sar}^{(k,l)} \right\}$ needs to consider shadowing effect. The latter is calculated using \autoref{equ_21}.

\begin{algorithm}[t]
	\small
	\renewcommand{\algorithmicrequire}{\textbf{Input:}}
	\renewcommand{\algorithmicensure}{\textbf{Output:}}
	\begin{algorithmic}[1]
		\STATE Initialization: $\left\{ \delta_j^{(k,l)} \right\}, \left\{ I_{sar}^{(k,l)} \right\}, \left\{ I_{sil}^{(k,l)} \right\}$
		\FOR {$l \leftarrow 1$ to $N_x$ and $k \leftarrow 1$ to $N_z$}
		\STATE $i_{sil} \leftarrow 1.$
		\STATE
		\FOR {$j \leftarrow 1$ to $N_f$}
		\STATE Calculate $\delta_j^{(k,l)}$ based on Eq. (6)
		\STATE $i_{sil} \leftarrow i_{sil} \cdot \left( 1 - \delta_j^{(k,l)} \right)$
		\ENDFOR
		\STATE 
		\FOR {$i \leftarrow 1$ to $N_y$}
		\STATE $s \leftarrow 0, i_{sar} \leftarrow 0$
		\FOR {$j \leftarrow 1$ to $N_f$}
		\STATE Calculate $\delta_j ^ {(i,l)}$ based on Eq. (6)
		\STATE Calculate $z_{j}^{(i,l)}$ based on Eq. (11)
		\STATE Calculate $d_z$ based on Eq. (9)
		\STATE $\rho_j ^ {(i,l)} \leftarrow \delta_j ^ {(i,l)} \exp \left( z_j ^ {(i,l)} / \gamma \right)$
		\STATE $s \leftarrow s + \rho_j ^ {(i,l)}$
		\STATE $i_{sar} \leftarrow i_{sar} + \delta_j^{(k,l)} \cdot S_j \cdot \rho_j^{(i,l)} \cdot \exp \left( - d_z^2 / 2 \sigma_{g} ^ 2 \right)$
		\ENDFOR
		\STATE $I_{sar}^{(k,l)} \leftarrow I_{sar}^{(k,l)} + i_{sar} / s$
		\ENDFOR
		\STATE
		\STATE $I_{sar}^{(k,l)} \leftarrow I_{sar}^{(k,l)} / {\sqrt{2 \pi}\sigma_{g}}$
		\STATE $I_{sil}^{(k,l)} \leftarrow 1-i_{sil}$
		\ENDFOR
		\ENSURE Rendered images $\left\{ I_{sar}^{(k,l)} \right\}, \left\{ I_{sil}^{(k,l)} \right\}$
	\end{algorithmic}	
	\label{alg_1}
	\caption{SAR rendering.}
\end{algorithm}

\section{Inverse Reconstruction}
With the proposed DSR, we equivalently establish a differentiable function between the input variables and the output rendering images. Using the error back propagation (BP) algorithm, we can back-propagate the difference between the output rendered image and the ground truth to the input, so as to learn the unknown input geometries and properties. After multiple iterations, when the output image converges with the ground truth, the 3D geometry is estimated, which means that 3D target reconstruction from the given images is accomplished. BP algorithm calculates the gradient of the output error to the input, and when calculating the gradient, the backward computation graph needs to be derived, and it has almost the same topology as the forward rendering framework.

\begin{figure}[!t]
	\centering
	\includegraphics[height=1.6in]{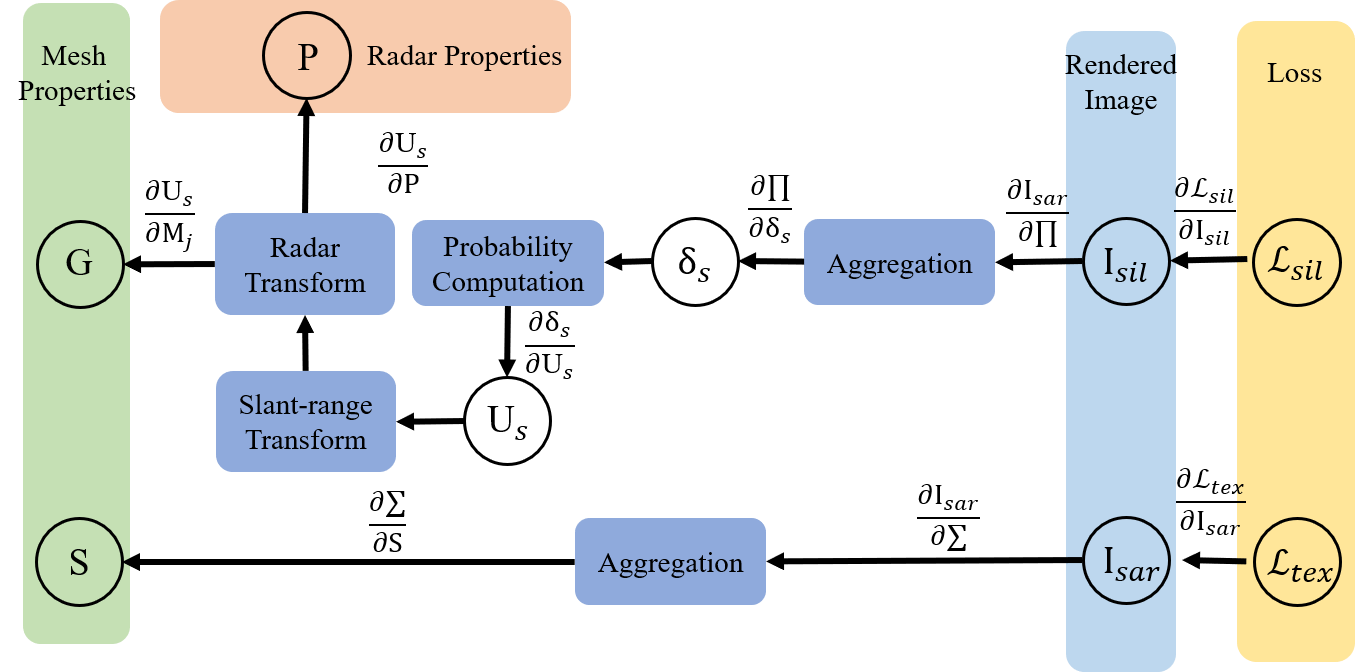}	
	\caption{Reconstruction framework.}
	\label{figure27}
\end{figure}

Like \autoref{figure1}, \autoref{figure27} provides the reverse gradient flow from the rendered image to the input variables and internal attributes. It is found that compared with \autoref{figure1}, several nodes, such as \textbf{Z}, \textbf{U} and \bm{${\rm \delta}$}, are missing in \autoref{figure27}, and the gradients of \bm{${\rm \delta_s}$}, \bm{${\rm \delta}$} and \textbf{Z}, which \bm{${\rm \mathbf{I}_{sar}}$} depends on, are not given. Considering that the rendering of \bm{${\rm \mathbf{I}_{sar}}$} involves too many variables, we render \bm{${\rm \mathbf{I}_{sil}}$} without using the depth and the scattering value in the previous section. The gradient calculation of \bm{${\rm \mathbf{I}_{sil}}$} to \bm{${\rm \mathbf{M}_{j}}$} is simpler and faster. Since BP algorithm is an algorithm that uses the chain rule to calculate differentiation, the gradient of a node to any adjacent node is calculated and marked on the edge between them. The aggregation operations of \autoref{equ_19} and \autoref{equ_20} are respectively summation and multiplication functions, so they are represented by \bm{${\rm \sum}$} and \bm{${\rm \prod}$}. When a node multiplies the returned gradient by the local gradient to its input, the gradient of the output error to each input of the node is obtained. After a recurrence from the end node of the renderer to the input one, the gradient of the output to the input is achieved. The gradient decent algorithm is used to adjust the mesh coordinate \textbf{G}, the facet scattering value \textbf{S} and the radar configuration \textbf{P} in each iteration. In this paper, 3D reconstruction is done when the viewing angles of the renderings are known, so \textbf{P} is a constant and no update is required.

\subsection{Gradient Flow}
This section derives the derivatives of the rendered images to the unknown parameters of target mesh. 3D structure can be inferred by multiple iterations of gradient decent algorithm. The initial guess is set to a sphere, textured with scattering value of $1$.

\subsubsection{Geometry Reconstruction}
Compared with $I_{sar}^{(k,l)}$, $I_{sil}^{(k,l)}$ does not consider the depth and the texture, so that there is no huge amount of calculation brought by energy calculating. $\delta_{j}^{(k,l)}$ is an intermediate variable connecting the silhouette $I_{sil}^{(k,l)}$ and the coordinate $\mathbf{M}_j$, so the derivative of $I_{sil}^{(k,l)}$ to $\mathbf{M}_{j}$ follows

\begin{equation}
	\label{equ_22}
	\frac{\partial I_{sil}^{(k,l)}}{\partial \mathbf{M}_{j}}=\frac{\partial I_{sil}^{(k,l)}}{\partial \delta_{j}^{(k,l)}} \frac{\partial \delta_{j}^{(k,l)}}{\partial \mathbf{M}_{j}}
\end{equation}

\subsubsection{Texture Reconstruction}
Each facet has only one texture, which directly participates in rendering the SAR image. According to \autoref{equ_19}, the gradient of the scattering value $S_{j}$ of $\mathbf{f}_j$ is
\begin{equation}
	\label{equ_26}
	\frac{\partial I_{sar}^{(k,l)}}{\partial S_{j}}=\delta_{j}^{(k,l)} \omega_{j}^{(k,l)}
\end{equation}

\subsection{Loss Function}
The reconstruction is supervised by a hybrid loss function. It is proposed for measuring the error between predicted images and the ground truths, i.e. 

\begin{equation}
	\label{equ_27}
	\mathcal{L}=\mathcal{L}_{s i l}+\lambda_{1} \mathcal{L}_{t e x}+\lambda_{2} \mathcal{L}_{l a p}+\lambda_{3} \mathcal{L}_{flat}
\end{equation}
where $\lambda_{1}=1$, $\lambda_{2}=0.03$, $\lambda_{3}=0.003$. The weight value decays as the importance decreases. $\mathcal{L}_{sil}$, $\mathcal{L}_{tex}$ denote the difference between the ground truth and the predicted silhouette and scattering images respectively. $\mathcal{L}_{sil}$ is negative intersection over union (IoU) between $\mathbf{I}_{sil}$ and $\hat{\mathbf{I}}_{sil}$, denoting the ground truth and the predicted silhouette respectively. 

\begin{equation}
	\label{equ_28}
	\mathcal{L}_{sil}=1-\frac{\mathbf{I}_{sil} \odot \hat{\mathbf{I}}_{sil}}{\mathbf{I}_{sil}+\hat{\mathbf{I}}_{sil}-\mathbf{I}_{sil} \odot \hat{\mathbf{I}}_{sil}}
\end{equation}
where $\odot$ denotes element-wise product.

$\mathcal{L}_{tex}$ is L1-norm between $\mathbf{I}_{sar}$ and $\hat{\mathbf{I}}_{sar}$.

\begin{equation}
	\label{equ_29}
	\mathcal{L}_{t e x}=\left\|\mathbf{I}_{sar}-\hat{\mathbf{I}}_{sar}\right\|_{1}
\end{equation}

$\mathcal{L}_{lap}$ adopts a random walk normalized Laplacian matrix $\mathbf{L}$. The element of the $i$-th row and $j$-th column in $\mathbf{L}$  follows

\begin{equation}
	\label{equ_30}
	L_{i j}=\left\{\begin{array}{c}
	1 \qquad  \text{if} \, i=j \, \text{and} \, \text{deg}\left(\mathbf{v}_{i}\right) \neq 0  \\
	-\frac{1}{\text{deg}\left(\mathbf{v}_{i}\right)} \quad \text{if} \, i \neq j \, \text{and} \, \mathbf{v}_{i} \, \text{is} \, \text{adjacent} \, \text{to} \, \mathbf{v}_{j} \\
	\ 0 \qquad  \text{otherwise} \qquad \qquad \qquad \qquad
	\end{array}\right.
\end{equation}
where $\text{deg}\left(\mathbf{v}_i\right)$ denotes the number of vertices adjacent to the vertex $\mathbf{v}_i$, named as degree of vertex.

In an image, Laplacian operation describes the difference between the central pixel and the local upper, lower, left and right neighbor pixels, and is usually used as an edge detection operator. In the same way, the operator can also be used to describe the signal difference between the central node and adjacent nodes in a mesh model, regarded as a graph. So the vertex set $\mathbf{V}$ is transformed to $\hat{\mathbf{V}}$, following

\begin{equation}
	\label{equ_31}
	\begin{gathered}
	\hat{\mathbf{V}} =\mathbf{L} \mathbf{V} = 
	\left[\cdots, \frac{1}{ \text{deg}\left(\mathbf{v}_{i}\right) } \sum_{\mathbf{v}_{j} \in N\left(\mathbf{v}_{i}\right)}\left( \mathbf{v}_{i} - \mathbf{v}_{j} \right), \cdots\right]^{\mathbf{T}}
	\end{gathered}
\end{equation}
where $N(\mathbf{v}_i)$ denotes the neighborhood of the vertex $\mathbf{v}_i$.

$\mathcal{L}_{lap}$ is the sum of squared coordinates in the Laplacian transform domain, as follows

\begin{equation}
	\label{32}
	\mathcal{L}_{lap} = \sum\limits_{i=1}^{N_v}{\sum\limits_{n=1}^{3}{(\hat{V}_{i,n})^{2}}}
\end{equation}

Assume that the two facets $f_1:\left\{ {\mathbf{v}_{1},\mathbf{v}_{2},\mathbf{v}_{3} }\right\}$ and $f_2:\left\{ {\mathbf{v}_{1},\mathbf{v}_{2},\mathbf{v}_{4} }\right\}$ share an edge. Project $\mathbf{v}_{3}$ and $\mathbf{v}_{4}$ to the edge $\left( \mathbf{v}_{2} - \mathbf{v}_{1} \right)$ and get the vertices $\mathbf{v}_{5}$ and $\mathbf{v}_{6}$. The angle $\theta$ between the two edges $\left( \mathbf{v}_{3} - \mathbf{v}_{5} \right)$ and $\left( \mathbf{v}_{4} - \mathbf{v}_{6} \right)$ is defined as

\begin{equation}
	\label{equ_33}
	\cos \theta=\frac{\left( \mathbf{v}_{3} - \mathbf{v}_{5} \right) \cdot \left( \mathbf{v}_{4} - \mathbf{v}_{6} \right) }{\left| \mathbf{v}_{3} - \mathbf{v}_{5} \right| \cdot \left| \mathbf{v}_{4} - \mathbf{v}_{6} \right|}
\end{equation}

Then $\mathcal{L}_{flat}$ follows

\begin{equation}
	\label{equ_34}
	\mathcal{L}_{flat}=\sum_{\theta_{i} \in \mathcal{E}}\left(1+\cos \theta_{i}\right)^{2}
\end{equation}
where $\mathcal{E}$ is the set of all edges in the deformed mesh \cite{IEEEhowto:Neural3D}. The purpose of decreasing $\mathcal{L}_{flat}$ is to make as more facets as possible co-planar, so that the mesh becomes smoother.

\subsection{Reconstruction Algorithm and Implement}
The pseudo code of the reconstruction is listed in Algorithm 2. It is mainly divided into two parts, which respectively returns the gradient tensor $\partial \mathbf{M}$ of vertex coordinates and the gradient tensor $\partial \mathbf{S}$ of facet's scattering value. Note that without the shadowing in the depth direction, the calculation of $\partial \mathbf{M}$ is much simpler than $\partial \mathbf{S}$. For each facet, according to \autoref{equ_22}, $\partial \mathbf{M}_{j}$ relates to $\delta_j ^{(k,l)}$; comparing \autoref{equ_19} and \autoref{equ_26}, it is found that the calculation of $\partial S_{j}$ is only one less iteration of the facet set $\left\{ \mathbf{f}_j \right\}$ than that of rendering the SAR image. It is necessary to accumulate the derivatives of all mapping cells to $\mathbf{M}_{j}$ and $S_{j}$ with respect to each facet $\mathbf{f}_j$.

Regardless of forward rendering or backward inversion, the entire renderer needs to calculate the interaction between each cell in both the projection and imaging planes and all the triangular facets. Matrix operations will require a huge memory and there is no guarantee that time consumption can be tolerated. Considering that these pixels are independent of each other, they can be regarded as multiple independent threads, which fits well the parallel computing scheme of Compute Unified Device Architecture (CUDA). Therefore, during the implementation based on the PyTorch framework, CUDA program is used to accelerate the calculation of probability maps and depth aggregation in the forward and backward programs.

\begin{algorithm}[h]
	\small
	\renewcommand{\algorithmicrequire}{\textbf{Input:}}
	\renewcommand{\algorithmicensure}{\textbf{Output:}}
	\begin{algorithmic}[1]
		\REQUIRE $\left\{ T^{(i,l)} \right\}, \left\{ \partial I^{(k,l)}_{sar} \right\}, \left\{ \partial I^{(k,l)}_{sil} \right\}, \left\{ I^{(k,l)}_{sil} \right\}$
		\STATE \% gradients of geometric coordinates and textures
		\STATE Initialization: $\left\{ \partial \mathbf{M}_{j} \right\}, \left\{ \partial S_j \right\}$
		\FOR {$l \leftarrow 1$ to $N_x$ and $k \leftarrow 1$ to $N_z$}
		\STATE \% iterate facets
		\FOR {$j \leftarrow 1$ to $N_f$}
		\STATE Calculate $\delta_j ^{(k,l)}$ based on Eq. (6)
		\STATE Update $\partial \mathbf{M}_{j}$ based on Eq. (22)
		\STATE
		\STATE $\rho \leftarrow 0$
		\FOR {$i \leftarrow 1$ to $N_y$}
		\STATE Calculate $\delta_j ^ {(i,l)}$ based on Eq. (6)
		\STATE Calculate $z_{j}^{(i,l)}$ based on Eq. (11)
		\STATE Calculate $d_z$ based on Eq. (9)
		\STATE $\rho_j ^ {(i,l)} \leftarrow \delta_j ^ {(i,l)} \exp \left( z_j ^ {(i,l)} / \gamma \right) / T^{(i,l)}$
		\STATE $\rho \leftarrow \rho + \rho_j ^ {(i,l)} \cdot \exp \left( d_z^{2} / 2 \sigma_{g} ^ 2 \right)$
		\ENDFOR
		\STATE $\partial S_j \leftarrow \partial S_j + \delta_j ^ {(k,l)} \cdot \rho \cdot \partial I_{sar}^{(k,l)} / \sqrt{2 \pi} \sigma_{g}$
		\STATE
		\ENDFOR
		\ENDFOR
		\ENSURE Gradients $\left\{ \partial \mathbf{M}_{j} \right\}, \left\{ \partial S_j \right\}$.
	\end{algorithmic}	
	\label{alg_2}
	\caption{Target reconstruction.}
\end{algorithm}

\section{Experiment}

\subsection{Target Rendering}
\begin{figure}[!t]
	\centering
	\subfloat[Building model $a$]{
		\includegraphics[width=1.6in]{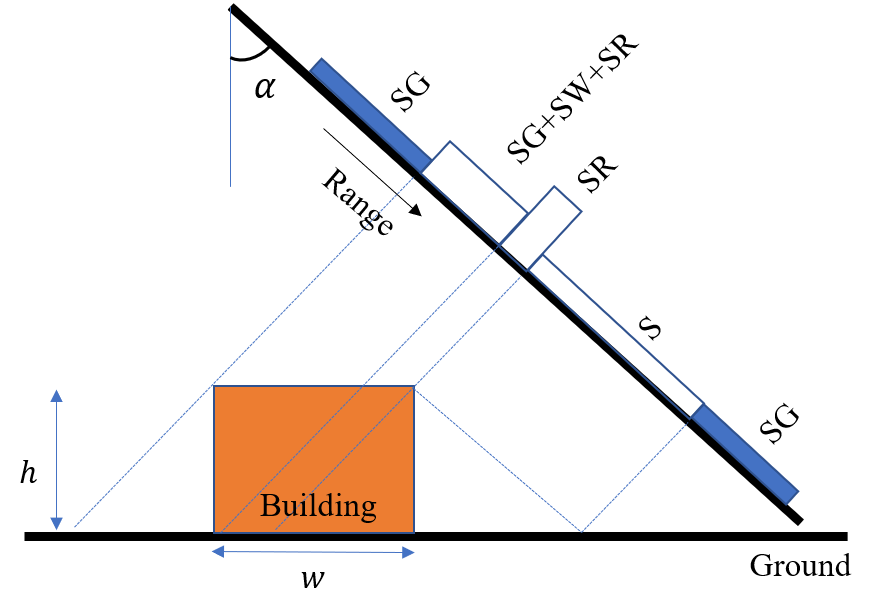}
	}
	\subfloat[The corresponding SAR image]{
		\includegraphics[width=1.6in]{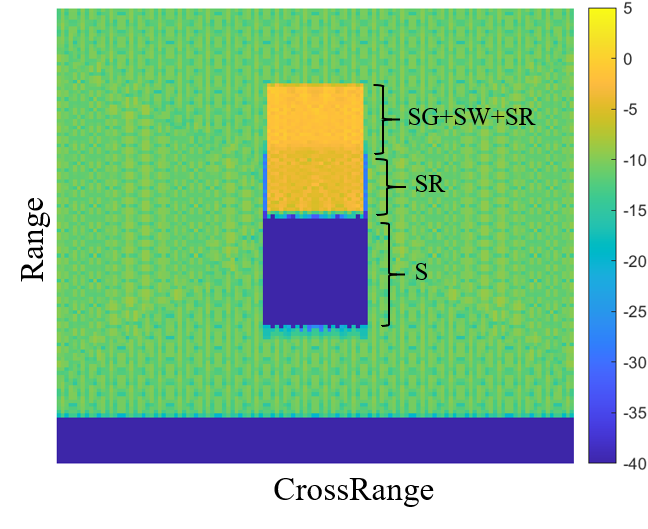}
	}
	\quad
	\subfloat[Building model $b$]{
		\includegraphics[width=1.6in]{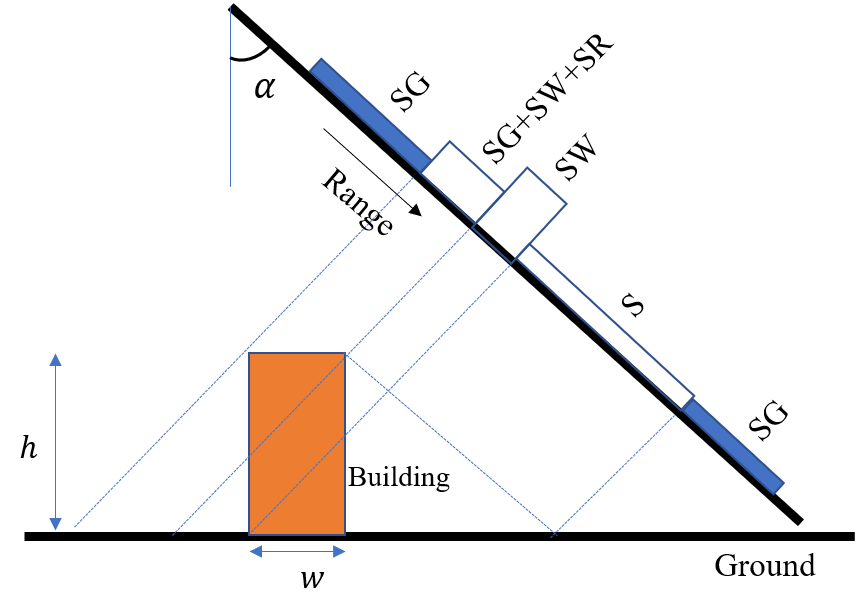}
	}
	\subfloat[The corresponding SAR image]{
		\includegraphics[width=1.6in]{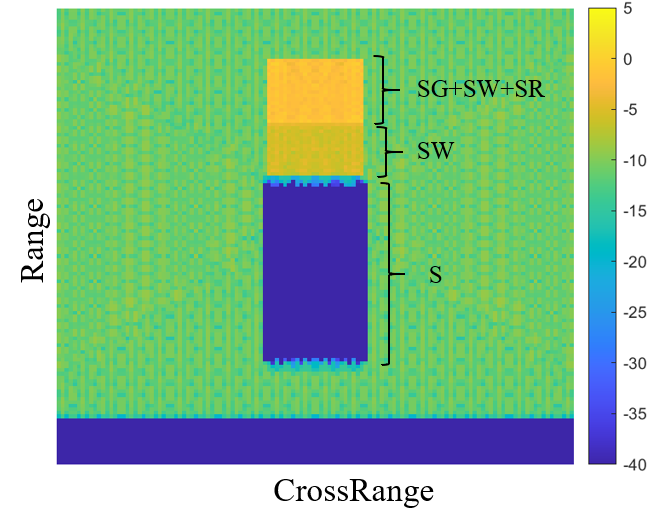}
	}
	\caption{Two kinds of building models and their corresponding rendered SAR images.}
	\label{figure7}
\end{figure}

We firstly verify the validity of results rendered by DSR. For a simple flat-topped building, it can be simplified as a cuboid. As is well known \cite{IEEEhowto:MPA}, a building as appeared in SAR image is consisted of ground scattering (SG), wall scattering (SW), roof scattering (SR) and shadow (S). The length of the shadow is proportional to the height of building. The scattering components distribution may vary as the height of wall $h$ and the width of roof $w$ vary. For the building model $a$ shown in \autoref{figure7}(a), it satisfies

\begin{equation}
	\label{equ_35}
	\frac{w}{h}>\cot \alpha 
\end{equation}
At this time, due to the large roof width, the area next to the SG+SW+SR layover is roof scattering. We increase the wall height so that it satisfies

\begin{equation}
	\label{equ_36}
	\frac{w}{h} < \cot \alpha 
\end{equation}
It corresponds to the model $b$ shown in \autoref{figure7}(c). Then the area after the layover is wall scattering. 

Here set $\alpha = 45 ^ \circ$, $w/h = 2$ for model $a$, and $w/h = 0.5$ for model $b$. In order to distinguish different scattering components, set different scattering intensities, 0.1, 0.5, 1.0, respectively for ground, wall and roof. Comparing \autoref{figure7}(b) and (d), it is found that the boundaries of different regions are distinct, and compositions of the building images rendered using DSR are correct. Note that all SAR images presented in this paper are in $10\log$ dB scale.

\subsection{Target Reconstruction with Rendered Images}
\begin{figure}[!t]
	\centering
	\subfloat[]{
		\includegraphics[height=1.0in]{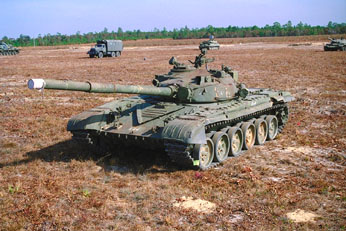}
	}
	\subfloat[]{
		\includegraphics[height=1.0in]{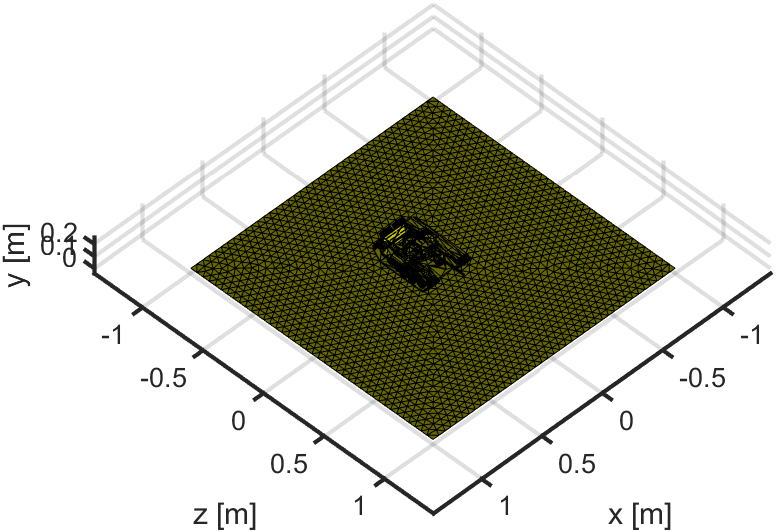}
	}
	\caption{CAD model of T72 vehicle. (a) Photograph of MSTAR T72 (from MSTAR dataset \cite{IEEEhowto:MSTAR}). (b) T72 vehicle mesh with ground.}
	\label{figure11}
\end{figure}

This section demonstrates ground target reconstruction with rendered images. The case used here is from the moving and stationary target acquisition (MSTAR) dataset \cite{IEEEhowto:MSTAR} and the T72 vehicle is tested. The 3D model of T72 is shown in \autoref{figure11} along with a picture of the actual setup.

\begin{figure}[!t]
	\centering
	\subfloat[T72 ground truth]{
		\includegraphics[width=1.6in]{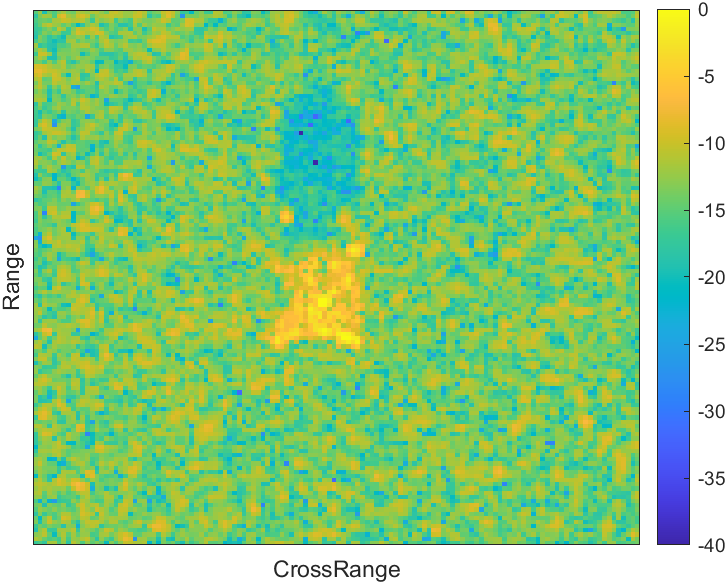}
	}
	\subfloat[Fit the distributions]{
		\includegraphics[width=1.6in]{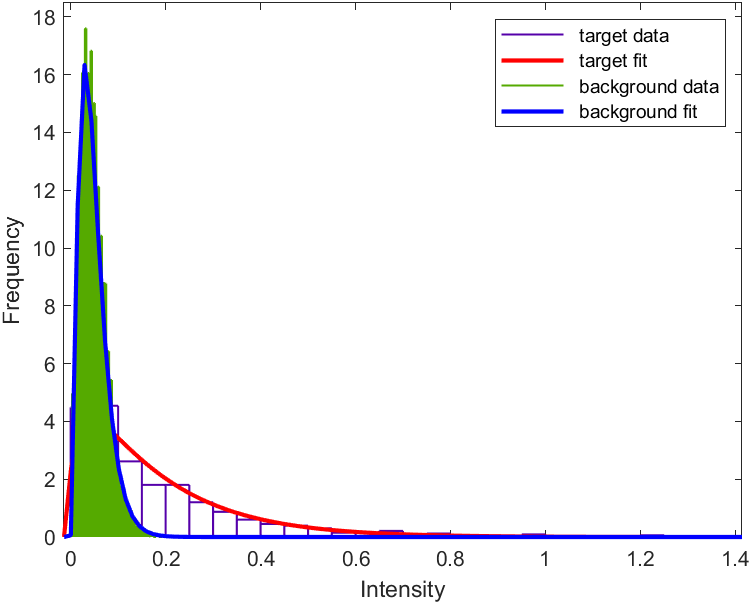}
	}
	\quad
	\subfloat[$\gamma = 1e-5$]{
		\includegraphics[width=1.6in]{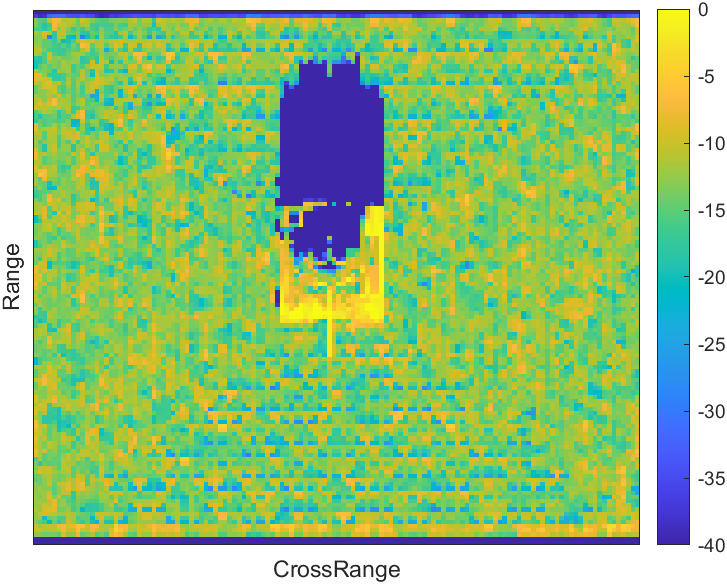}
	}
	\subfloat[$\gamma = 1e-2$]{
		\includegraphics[width=1.6in]{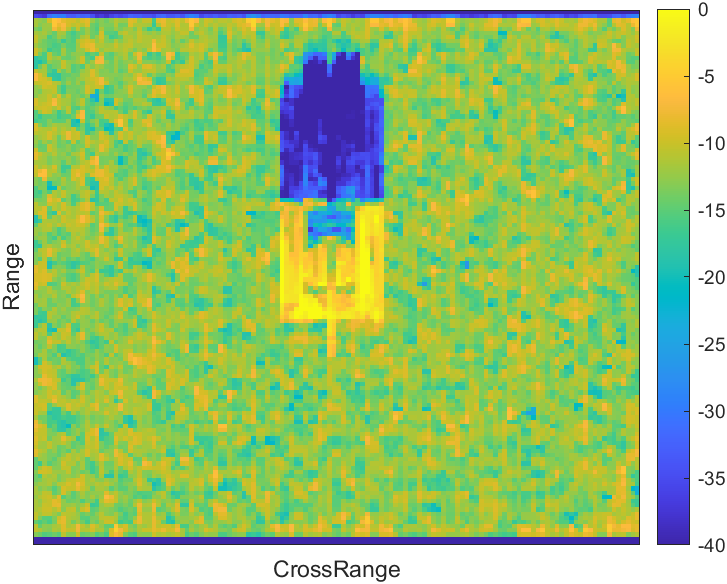}
	}
	\caption{Rendering images for the T72 vehicle.}
	\label{figure12}
\end{figure}

In order to have realistic scattering textures of the model, we first take a real MSTAR image (\autoref{figure12}(a)) and fit Gamma distributions to the target area and ground area (\autoref{figure12}(b)). The fitted hyper-parameters are listed in \autoref{table1}. Subsequently, we generate random samples from the Gamma distributions and respectively assign to the scattering textures of the facets of the target and the ground surface. Finally, we feed the model into the DSR, and obtain the rendered image at viewing angle of $\alpha = 75 ^ \circ$, $\beta = 0 ^ \circ$ (\autoref{figure12}(c,d)).

\begin{table}[!t]
	\caption{Gamma distribution parameters for texture fitting of target and background.}
	\label{table1} 
	
	\renewcommand{\arraystretch}{1.3}
	\setlength{\tabcolsep}{4mm}
	\centering
	
	\begin{tabular}{c c c}
		\hline \hline
		& \textbf{Shape Parameter} & \textbf{Scale Parameter} \\
		\hline
		Target & 1.1948 & 0.1508 \\
		\hline
		Background & 2.7179 & 0.0177 \\
		\hline \hline
	\end{tabular}
	
\end{table}

We first set $\sigma=1e-5$, $\gamma=1e-5$, $\sigma_g=0.5$. When $\gamma \to 0$, the facets facing the radar will completely obscure the facets facing away from the radar, which corresponds to depth buffering of the traditional renderer. When the incident angle $\alpha = 75 ^ \circ$ is large, facets at the tail are easily blocked by facets at the front. It results in many pixels of the tail recognized as shadow when $\gamma = 1e-5$ in \autoref{figure12}(c). When $\gamma = 1e-2$ in \autoref{figure12}(d), facets facing away from the radar become 'partially visible', and facets at the tail will show up in the image.

\begin{figure}[!t]
	\centering
	\subfloat[Ground truth SAR image]{
		\includegraphics[width=1.6in]{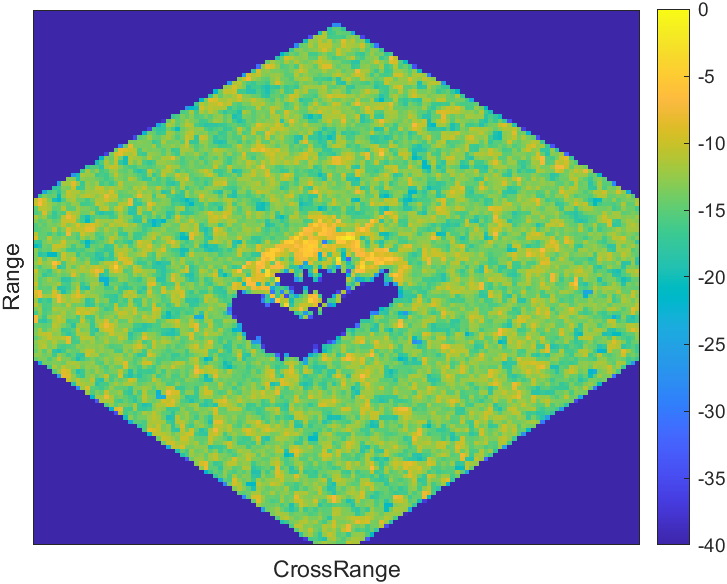}
	}
	\subfloat[Segmented silhouette of T72 vehicle]{
		\includegraphics[width=1.6in]{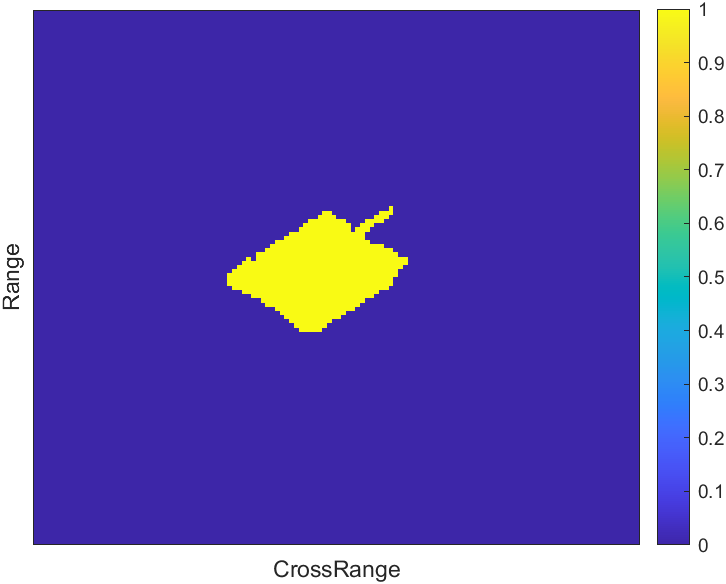}
	}
	\quad
	\subfloat[Segmented SAR image of T72 vehicle]{
		\includegraphics[width=1.6in]{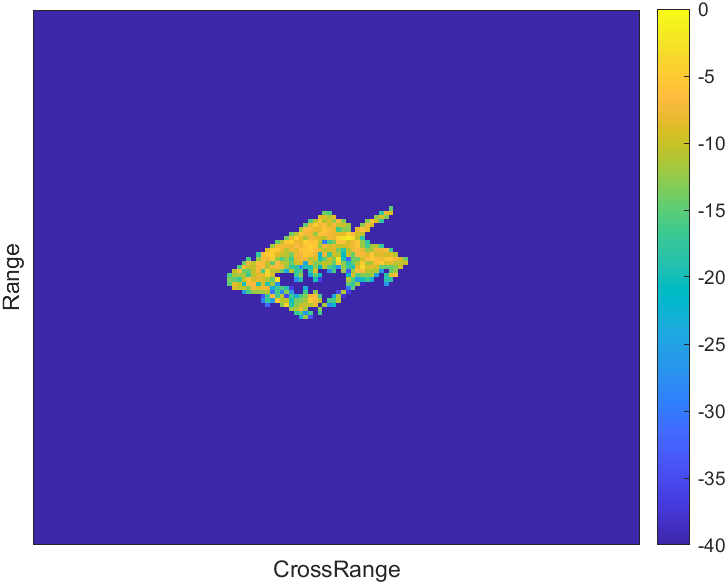}
	}
	\subfloat[Predicted SAR image]{
		\includegraphics[width=1.6in]{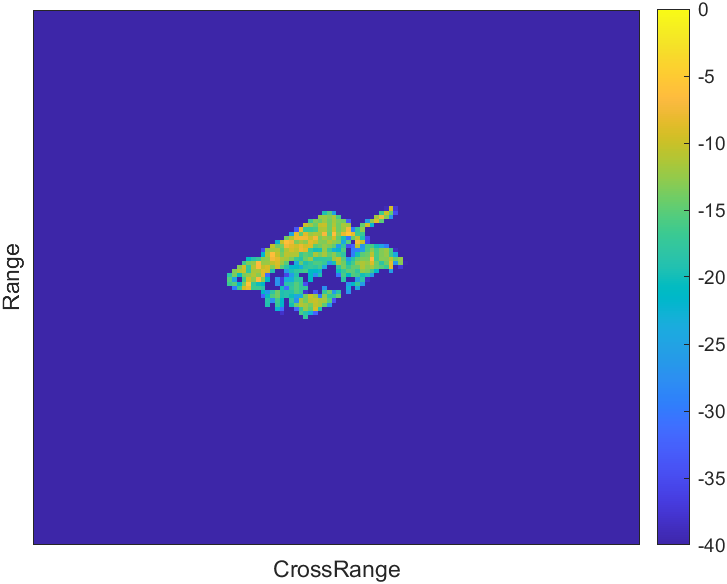}
	}
	\caption{Images at $\alpha=45^\circ, \beta=45^\circ$.}
	\label{figure14}
\end{figure}

We render SAR images from four different incident angles and eight different aspect angles. These angles are evenly distributed, i.e. $\alpha = \left\{15^\circ, 30^\circ, 45^\circ, 60^\circ \right\}$, $\beta = \left\{0^\circ, 45^\circ, 90^\circ, 135^\circ, 180^\circ, 225^\circ, 270^\circ, 315^\circ \right\}$, and finally we get a total of 32 images. For instance, \autoref{figure14}(a) is rendered at $\alpha = 45 ^ \circ$, $\beta = 45 ^ \circ$. 

To reverse the geometric structure, the silhouettes of the target should be provided. So firstly segment the target out of the ground background to a binary mask. The segmentation can be realized using image processing techniques as shown in \autoref{figure14}(b). The initialization of the renderer is a sphere mesh for reconstruction. A 3D structure can be inferred by performing BP.

The rendered image size is $128 \times 128$. The learning rate is 0.01 and stochastic gradient descent algorithm with adaptive moment estimation (Adam) is used. The batch size, abbreviated to $bs$, is set 8, and the number of epochs is 500. 

\begin{figure}[!t]
	\centering
	\subfloat[T72 ground truth]{
		\includegraphics[width=1.0in]{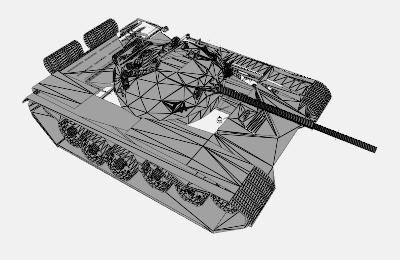}
	}
	\subfloat[$bs = 1$]{
		\includegraphics[width=1.0in]{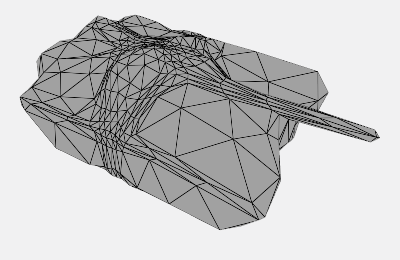}
	}
	\subfloat[$bs = 4$]{
		\includegraphics[width=1.0in]{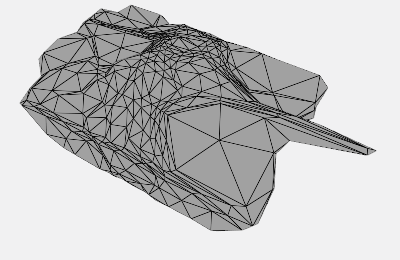}
	}
	\quad
	\subfloat[$bs = 8$]{
		\includegraphics[width=1.0in]{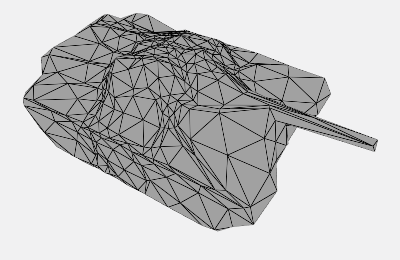}
	}
	\subfloat[no $\mathcal{L}_{flat}$]{
		\includegraphics[width=1.0in]{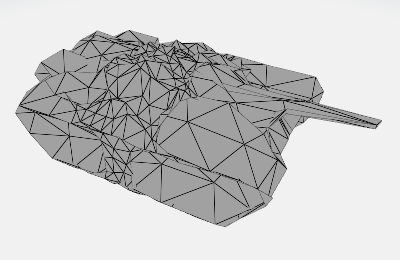}
	}
	\subfloat[no $\mathcal{L}_{lap}$]{
		\includegraphics[width=1.0in]{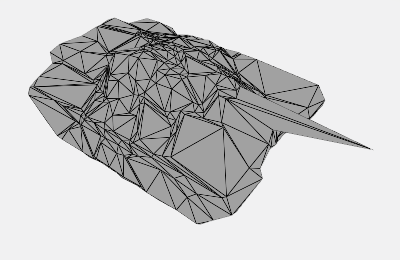}
	}
	\quad
	\subfloat[no $\mathcal{L}_{flat}$ \& $\mathcal{L}_{lap}$]{
		\includegraphics[width=1.0in]{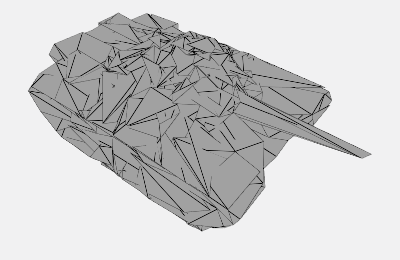}
	}
	\caption{Comparison between ground truth and the reconstructed meshes.}
	\label{figure15}
\end{figure}

When $bs$ is set to a large value, due to the large difference between a batch of samples, taking the average gradients will neutralize the negative impact of some biased samples. The geometries of 3D vehicles reconstructed with different batch sizes don't vary so much in \autoref{figure15}. Fix the batch size as $bs = 8$ and respectively remove $\mathcal{L}_{flat}$ and $\mathcal{L}_{lap}$ in the loss function. It is found that $\mathcal{L}_{flat}$ has a greater positive impact on the surface smoothness.

\begin{figure}[!t]
	\centering
	\subfloat[T72 ground truth]{
		\includegraphics[width=1.0in]{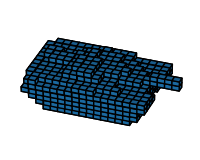}
	}
	\subfloat[$bs=1$]{
		\includegraphics[width=1.0in]{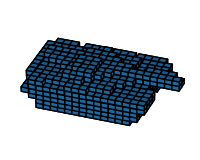}
	}
	\subfloat[$bs=4$]{
		\includegraphics[width=1.0in]{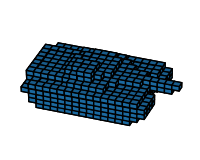}
	}
	\quad
	\subfloat[$bs=8$]{
		\includegraphics[width=1.0in]{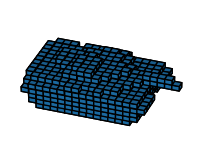}
	}
	\subfloat[no $\mathcal{L}_{flat}$]{
		\includegraphics[width=1.0in]{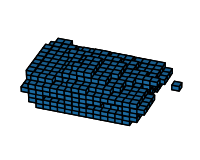}
	}
	\subfloat[no $\mathcal{L}_{lap}$]{
		\includegraphics[width=1.0in]{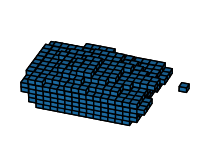}
	}
	\quad
	\subfloat[no $\mathcal{L}_{flat}$ \& $\mathcal{L}_{lap}$]{
		\includegraphics[width=1.0in]{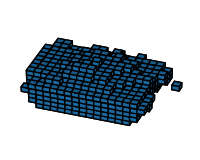}
	}
	\caption{Voxelization of meshes in \autoref{figure15}.}
	\label{figure28}
\end{figure}

To evaluate the reconstruction performance quantitatively, we voxelize the meshes reconstructed with different parameter settings \cite{IEEEhowto:Neural3D} in \autoref{figure28}. Then calculate the IoU between voxels. The definition of 3D IoU is similar to that of 2D IoU, both of which calculate the intersection over union of pixels/voxels. The size of voxels is set to $32 ^ 3$. For each target, we perform 3D reconstruction using the silhouettes from 32 viewpoints, calculate the IoU scores, and record the average score.

As shown in \autoref{table2}, as the batch size increases, the IoU score also increases. However, when we remove $\mathcal{L}_{lap}$ or $\mathcal{L}_{flat}$ from the loss function, their IoU scores are better than that iterated with the integral $\mathcal{L}$. Although the surfaces of meshes in \autoref{figure15}(e), (f) and (g) are coarse, this shortcoming disappears during voxelization. That is to say, the voxels-IoU cannot reflect the smoothness of the mesh.

\begin{table}[!t]
	\caption{Qualitative elevations under different parameter settings.}
	\label{table2} 
	
	\renewcommand{\arraystretch}{1.3}
	\setlength{\tabcolsep}{6mm}
	\centering
	
	\begin{tabular}{c c c c}
		\hline \hline
		\textbf{Batch Size} & $\mathcal{L}_{l a p}$ & $\mathcal{L}_{flat}$ & \textbf{mIoU} \\
		\hline
		1 & $\checkmark$ & $\checkmark$ & 0.5542 \\
		\hline
		4 & $\checkmark$ & $\checkmark$ & 0.5742 \\
		\hline
		8 & $\checkmark$ & $\checkmark$ & 0.5614 \\
		\hline
		8 & $\checkmark$ & & 0.5735 \\
		\hline
		8 & & $\checkmark$ & 0.5685 \\
		\hline 
		8 &  &  & 0.5807 \\
		\hline \hline
	\end{tabular}
	
\end{table} 

\subsection{Target Reconstruction with Simulated Images}
\begin{figure}[!t]
	\centering
	\subfloat[]{
		\includegraphics[height=1.4in]{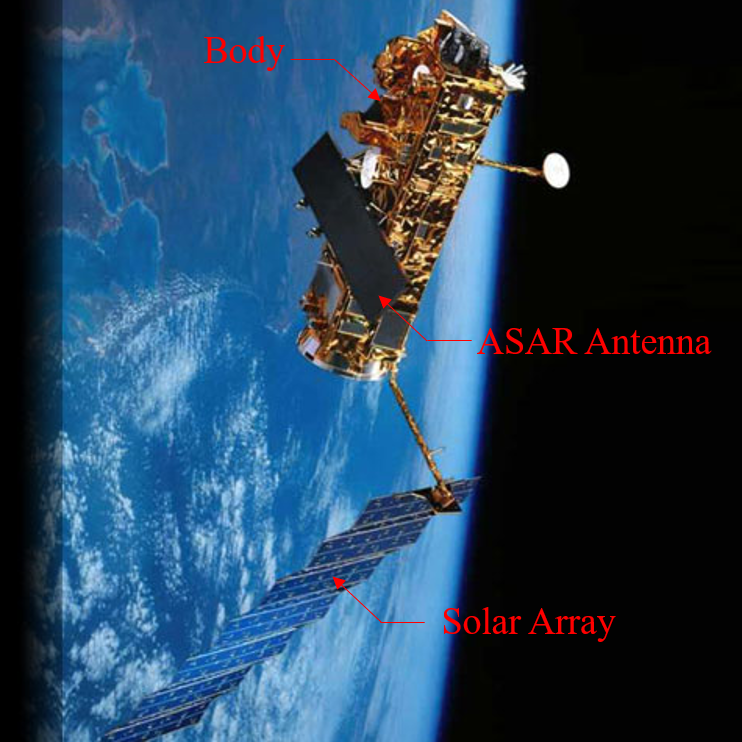}
	}
	\hspace{6mm}
	\subfloat[]{
		\includegraphics[height=1.4in]{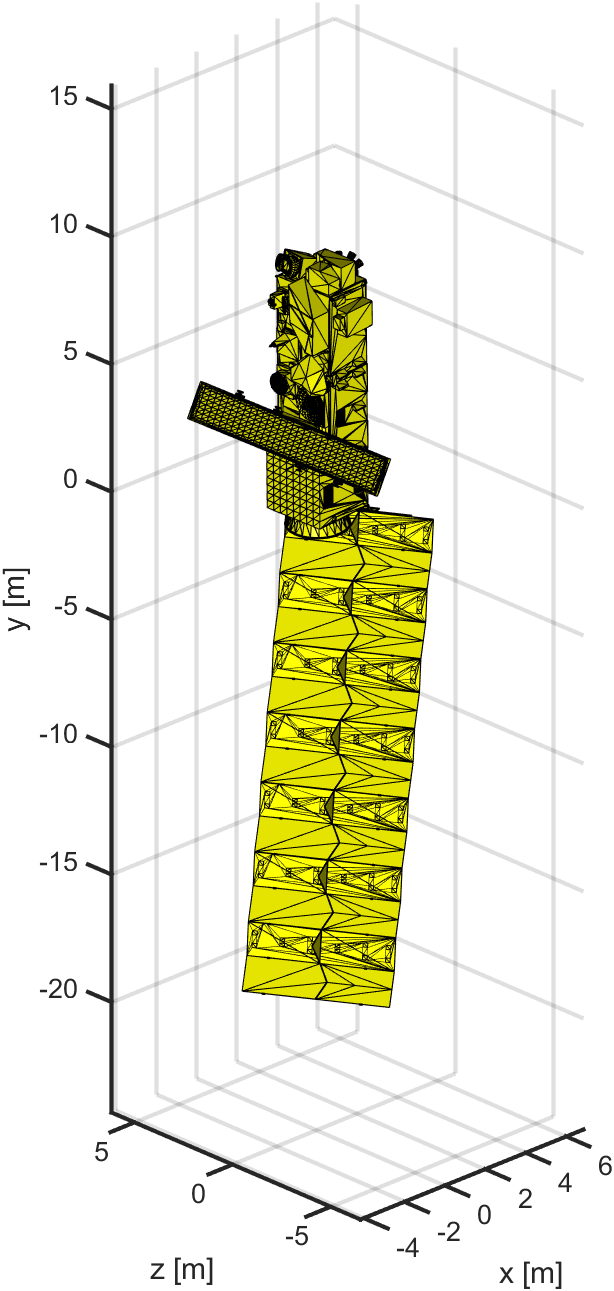}
	}
	\caption{(a) Photograph of the Envisat satellite (from Internet), (b) its CAD model.}
	\label{figure17}
\end{figure}

An attempt to reconstruct the target in the simulated images is made here. To avoid background interference, we now turn to high-resolution inverse SAR (ISAR) images of space targets. The Envisat satellite and its geometry model are shown in the \autoref{figure17}. The satellite consists of two rectangles (a solar array and an advanced SAR antenna), and approximately a cuboid (satellite body). 

We adopt bidirectional analytic ray tracing (BART) method \cite{IEEEhowto:BART} to calculate the scattering and obtain a multi-view ISAR imaging sequence. The incident angles are respectively set as $\left\{ -30^\circ, -40^\circ, -50^\circ, -60^\circ \right\}$, and the azimuth angle is sampled at intervals of $45^\circ$. The parameters are listed in \autoref{table3}. Its resolutions in the azimuth- and slant-directions are, respectively, given as
\begin{equation}
	\label{equ_37}
	r_a = \frac{c}{2 f_c \Delta \phi}, \quad r_r = \frac{c}{2 B}
\end{equation}
where $c$ is the speed of light, $f_c$ is the center frequency, $\Delta \phi$ is the angular bandwidth in the azimuth direction, and $B$ is the frequency bandwidth.

\begin{table}[!t]
	\caption{Parameters for ISAR  Imaging Simulation of the Envisat Model.}
	\label{table3} 
	
	\renewcommand{\arraystretch}{1.3}
	\setlength{\tabcolsep}{7mm}
	\centering
	
	\begin{tabular}{l l}
	\hline \hline
	\textbf{Parameter} & \textbf{Quantity} \\
	\hline
	Number of facets & 18391 \\
	$f_c$ (center frequency) & 16.7GHz \\
	$B$ (bandwidth) & 1GHz \\
	Number of frequencies & 251 \\
	Resolution in range direction & $0.150m$ \\
	Accumulation angle & $3.5 ^ \circ$ \\
	Number of angles & 251 \\
	Resolution in cross-range direction & $0.147m$ \\
	\hline \hline
	\end{tabular}
	
\end{table}

\begin{figure}[!t]
	\centering
	\subfloat[$\alpha=-30 ^ \circ, \beta=225 ^ \circ$]{
		\includegraphics[width=1.6in]{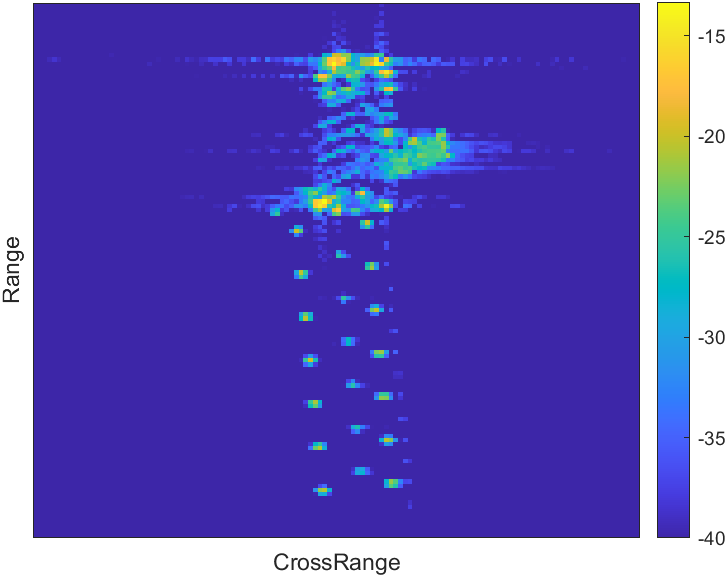}
	}
	\subfloat[$\alpha=-60 ^ \circ, \beta=225 ^ \circ$]{
		\includegraphics[width=1.6in]{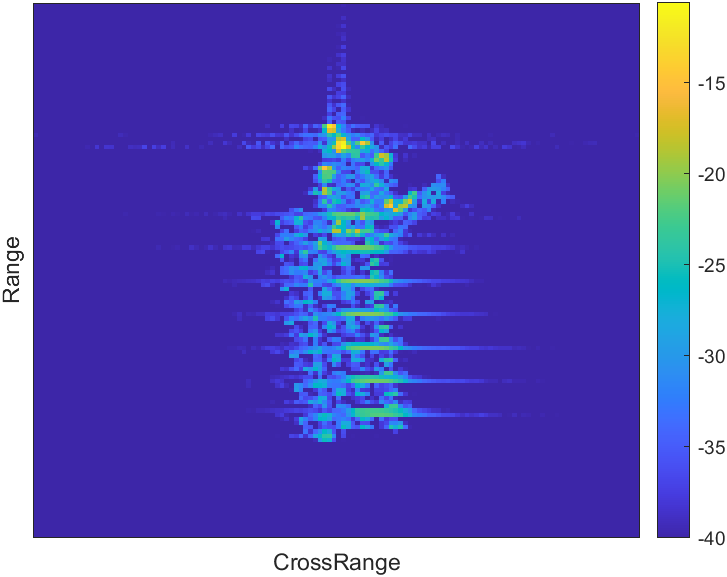}
	}
	\caption{Simulated images at different incident angles.}
	\label{figure18}
\end{figure}

The simulated images under different incident angles are shown in \autoref{figure18}. It seems that the simulated results have severe sidelobe effects, while this is not the case for images rendered by DSR which does not simulate the focusing effect. Apparently, such sidelobes will cause interference to the inverse reconstruction. It is necessary to filter the sidelobes. For an imaging system, a tiny triangular facet smaller than a resolution cell can be regarded as a point target, and the simulated image is equivalent to the point spread function (PSF). PSF can be used for filtering the sidelobes as

\begin{equation}
	\label{equ_39}
	\hat{x}\left(i^{\prime}, j^{\prime}\right)=\left\{\begin{array}{c}
	x\left(i^{\prime}, j^{\prime}\right)  \text { if } x\left(i^{\prime}, j^{\prime}\right)-x(i, j)> \text{PSF}(\Delta i, \Delta j) \\
	0  \qquad  \text { otherwise } \qquad \qquad \qquad \qquad \qquad \,
	\end{array}\right.
\end{equation}
where $\Delta i = \left|i' - i\right|, \Delta j = \left| j' - j \right|$.  $x, \hat{x}$ respectively denote SAR images before and after filtering. 

After removing sidelobes, the simulated images are mainly composed of scattering points, especially the solar array part with zigzags on the surface. Subsequently, we annotate the target's binary masks as ground-truth silhouettes.

\begin{figure}[!t]
	\centering
	\subfloat[Sidelobe suppression for \autoref{figure18}(b)]{
		\includegraphics[width=1.6in]{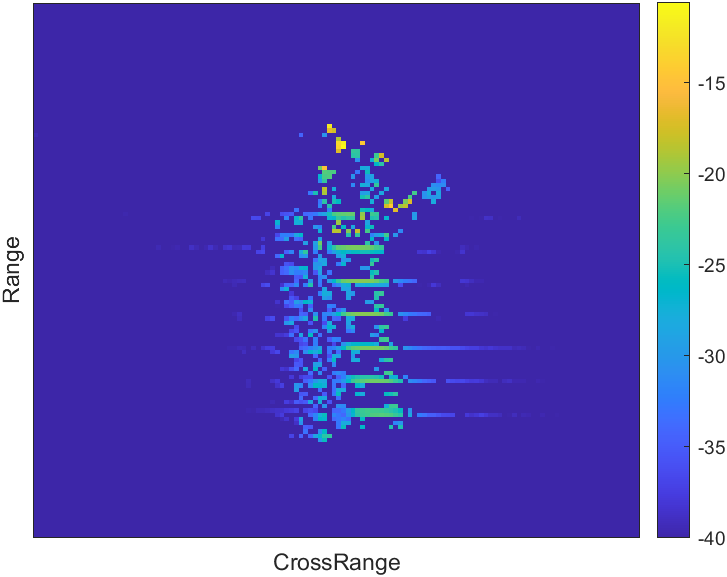}
	}
	\subfloat[Binary mask.]{
		\includegraphics[width=1.6in]{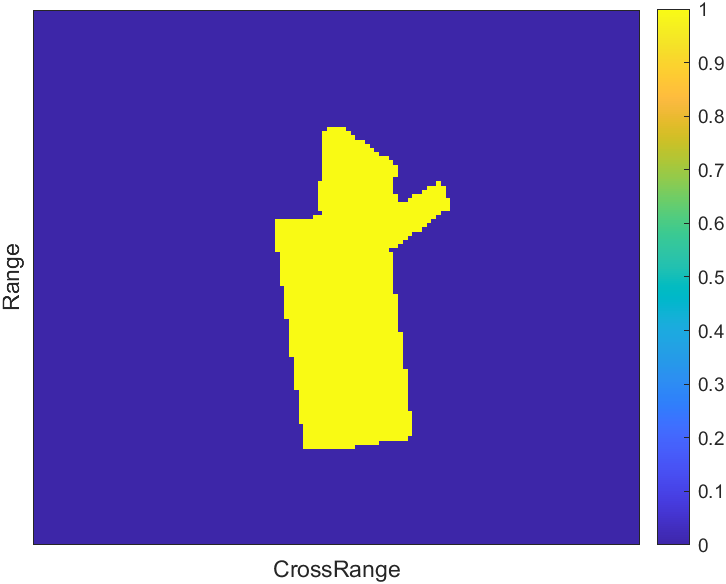}
	}
	\caption{Sidelobe suppression and manual annotating.}
	\label{figure20}
\end{figure}

\begin{figure}[!t]
	\centering
	\subfloat[Ground truth]{
		\includegraphics[width=1.4in]{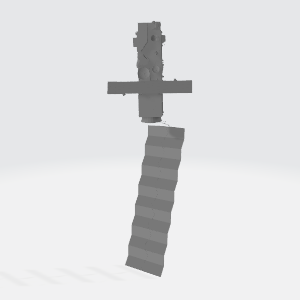}
	}
	\hspace{3mm}
	\subfloat[Reconstructed mesh]{
		\includegraphics[width=1.4in]{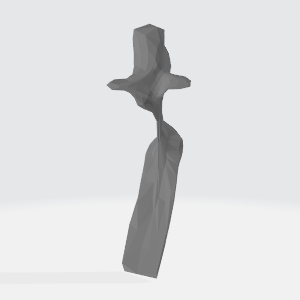}
	}
	\caption{Comparison between the ground truth satellite and the reconstructed mesh.}
	\label{figure21}
\end{figure}

After 200 iterations, the loss quickly converges to a small value, and a reconstructed 3D satellite model is obtained. The reconstructed satellite in \autoref{figure21}(b) is also composed of three parts, which accurately expresses the topological structure of the target. Compared to the ground truth, the solar array of the reconstructed satellite is almost flat. This is partly due to $\mathcal{L}_{flat}$, and partly because the panel part of the silhouettes provided to the renderer is rectangular.

\subsection{Target Reconstruction with Real ISAR Images}
\begin{figure}[!t]
	\centering
	\subfloat[Posture diagram]{
		\includegraphics[width=1.2in]{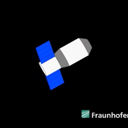}
	}
	\hspace{5mm}
	\subfloat[Heat map]{
		\includegraphics[width=1.2in]{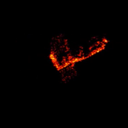}
	}
	\quad
	\subfloat[Binary map]{
		\includegraphics[width=1.2in]{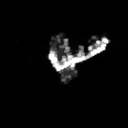}
	}
	\hspace{5mm}
	\subfloat[Annotated map]{
		\includegraphics[width=1.2in]{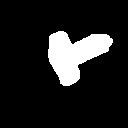}
	}
	\caption{Frame 240 in the video.}
	\label{figure22}
\end{figure}

We further investigate 3D reconstruction using a set of real ISAR images of the Tiangong-1 space station. Tiangong-1 mainly consists of a cylindrical cabin and a pair of long horizontal solar cell wings. FGAN laboratory \cite{IEEEhowto:flab} released a sequence of ISAR imaging results of Tiangong-1. Its original format is a video consisted of 396 frames. We extract some key frames and resize to $128 \times 128$ heat maps. They are further binarized to binary maps which are then annotated to ground-truth silhouettes as shown in \autoref{figure22}(b,c,d). 

The first 240 frames of video are rotating views of the Tiangong-1 around $O'Z'$ axis in the imaging coordinate system. This transformation cannot be achieved by changing the incident angle and the azimuth angle, so Euler angles $\left( \theta_x, \theta_y, \theta_z \right)$ are introduced to make the target capable of yaw, pitch and roll in the radar coordinate system. The transformation matrices of rotating $\theta$ around the axis $O'X'$, $O'Y'$, $O'Z'$ can be respectively denoted as


\begin{equation}
	\label{equ_40}
	\begin{gathered}
		{{R}_{x}}\left( {{\theta}} \right)=\left[ \begin{matrix}
			1 & 0 & 0  \\
			0 & \cos {{\theta}} & \sin {{\theta}}  \\
			0 & -\sin {{\theta}} & \cos {{\theta}}  \\
		\end{matrix} \right],
		{{R}_{y}}\left( {{\theta}} \right)=\left[ \begin{matrix}
			\cos {{\theta}} & 0 & -\sin {{\theta}}  \\
			0 & 1 & 0  \\
			\sin {{\theta}} & 0 & \cos {{\theta}}  \\
		\end{matrix} \right] \\ 
		{{R}_{z}}\left( {{\theta}} \right)=\left[ \begin{matrix}
			\cos {{\theta}} & \sin {{\theta}} & 0  \\
			-\sin {{\theta}} & \cos {{\theta}} & 0  \\
			0 & 0 & 1  \\
		\end{matrix} \right] \\ 
	\end{gathered}
\end{equation}

So Euler transformation with $\left( \theta_x, \theta_y, \theta_z \right)$ corresponds to rotating $\theta_x$, $\theta_y$, $\theta_z$ around the axis $O'X'$, $O'Y'$, $O'Z'$ respectively.

\begin{equation}
	\label{equ_41}
	{{R}_{e}}\left( {{\theta }_{x}},{{\theta }_{y}},{{\theta }_{z}} \right)={{R}_{y}}\left( {{\theta }_{y}} \right){{R}_{x}}\left( {{\theta }_{x}} \right){{R}_{z}}\left( {{\theta }_{z}} \right)
\end{equation}


A total of 9 key-frame silhouettes are selected. However, the angle corresponding to each frame is unknown. We estimate the viewing angles by mannually comparing the measured silhouettes to DSR rendered ones using the ground-truth Tiangong-1 model. The manually estimated viewing angles are shown in \autoref{table5}.

\begin{table}[!t]
	\caption{Estimated angles for the nine frames in the video.}
	\label{table5} 
	
	\renewcommand{\arraystretch}{1.5}
	\setlength{\tabcolsep}{1mm}
	\centering
	
	\begin{tabular}{p{1.5cm}<{\centering} p{1.5cm}<{\centering} p{1.5cm}<{\centering} p{1.5cm}<{\centering} p{1.5cm}<{\centering}}
		\hline \hline
		\textbf{Number} & \textbf{Frame Number} & \textbf{Incident Angle/$^\circ$} & \textbf{Azimuth Angle/$^\circ$} & \textbf{Euler Angle/$^\circ$} \\
		\hline
		1 & 30 & 75 & 0 & (0,0,135)  \\
		\hline
		2 & 60 & 75 & 0 & (0,0,180) \\
		\hline
		3 & 85 & 75 & 0 & (0,0,200) \\
		\hline
		4 & 160 & 75 & 0 & (0,0,0) \\
		\hline
		5 & 240 & 60 & 0 & (-10,0,90) \\
		\hline
		6 & 280 & -20 & 0 & (0,225,0) \\
		\hline
		7 & 320 & -20 & 0 & (0,180,0) \\
		\hline
		8 & 370 & -20 & 0 & (0,150,0) \\
		\hline
		9 & 390 & -20 & 0 & (0,135,0) \\
		\hline \hline
	\end{tabular}
	
\end{table}


Since the scattering intensity value of the target cannot be obtained from the heat map, the scattering values of the facets in the 3D model cannot be inferred. Only the geometric shape can be reconstructed here, so the loss function should be modified as

\begin{equation}
	\label{equ_43}
	\mathcal{L}=\mathcal{L}_{s i l} +\lambda_{2} \mathcal{L}_{l a p}+\lambda_{3} \mathcal{L}_{flat}
\end{equation}

\begin{figure}[!t]
	\centering
	\subfloat[DSR's silhouette]{
		\includegraphics[width=1.2in]{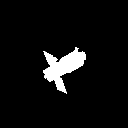}
	}
	\hspace{5mm}
	\subfloat[Mesh reconstructed with (a)]{
		\includegraphics[width=1.6in]{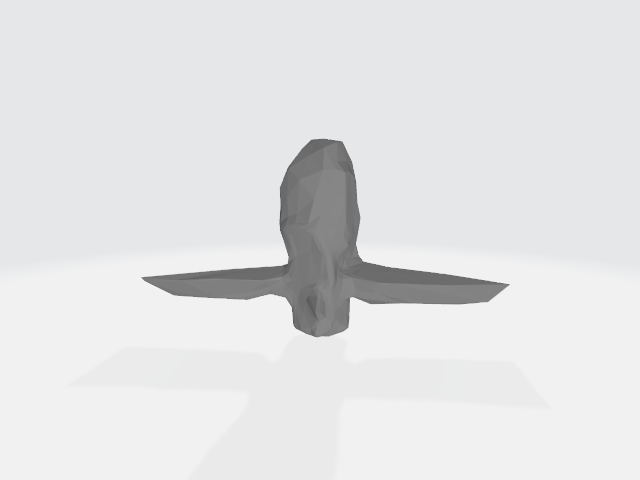}
	}
	\quad
	\subfloat[Video frame's silhouette]{
		\includegraphics[width=1.2in]{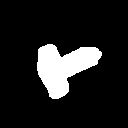}
	}
	\hspace{5mm}
	\subfloat[Mesh reconstructed with (c)]{
		\includegraphics[width=1.6in]{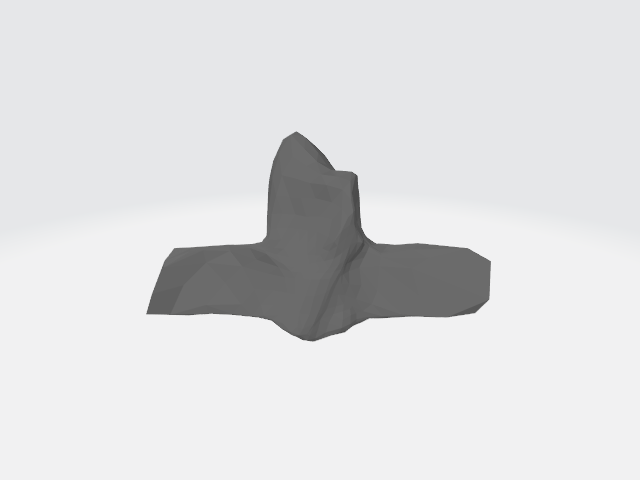}
	}
	\caption{Comparison between different reconstructed meshes.}
	\label{figure25}
\end{figure}

When the number of silhouette is small, the batch size is set to 9. After 500 iterations, the loss quickly converges. While reconstruct the model from the same 9 perspectives,

\begin{itemize}
	\item the model reconstructed with silhouettes rendered by DSR in \autoref{figure25}(b) is more slender and the tail of the cabin is more obvious;
	
	\item The model reconstructed with silhouettes from the video in \autoref{figure25}(d) is coarser. During the annotation process, it is found that the tail of the cabin is almost covered by solar cell wings, resulting in the tail not protruding enough.
\end{itemize}

\subsection{Target Attitude Estimation}
This section also considers using DSR to estimate the orientations. The inversion of the model geometry and mesh facet's texture has been realized under known viewing angles. Then in the case of a known model, it should be feasible to infer the viewing angles and thus the target orientation. Since the viewing angles are estimated based on the silhouette, we set $\mathcal{L}=\mathcal{L}_{s i l}$.

\begin{figure}[!t]
	\centering
	\subfloat[Ground truth silhouette]{
		\includegraphics[width=1.6in]{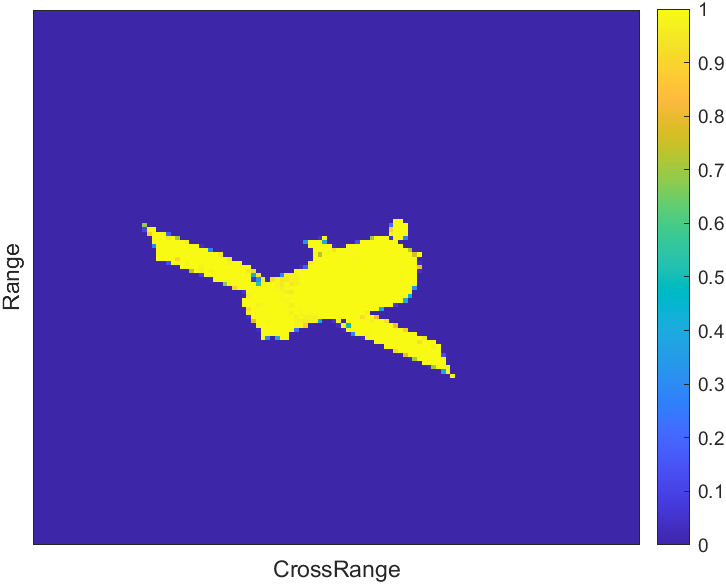}
	}
	\subfloat[Predicted silhouette]{
		\includegraphics[width=1.6in]{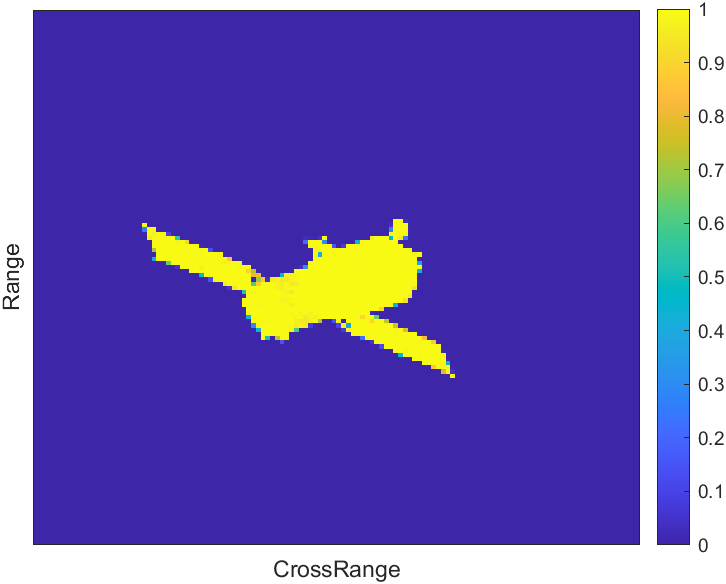}
	}
	\caption{Angle inversion.}
	\label{figure26}
\end{figure}

Taking the Tiangong-1 space station model as an example, first set the perspective of the renderer as the ground truth in \autoref{table8}, and get the silhouette as \autoref{figure26}(a). Initialize the values close to the ground truth to speed up the convergence, and after the iteration of 500 epochs, the predicted silhouette is almost the same as the ground truth. 

As shown in \autoref{table8}, it is found that the predicted angles don't match the ground truths. This is because Euler angles are not completely independent of the incident and azimuth angles. A change in Euler angles may be equivalent to another change in the incident and azimuth angles. The reason for introducing the scale factor here is that there may be an overall scaling of the model before rendering.

\begin{table}[!t]
	\caption{Comparison between ground truth and predicted angles.}
	\label{table8} 
	
	\renewcommand{\arraystretch}{1.8}
	\setlength{\tabcolsep}{1mm}
	\centering
	
	\begin{tabular}{c c c c}
		\hline \hline
		& \textbf{Ground Truth} & \textbf{Initialization} & \textbf{Prediction} \\
		\hline
		Incident angle/$^\circ$ & 75 & 60 & 66.5114 \\
		\hline
		Azimuth angle/$^\circ$ & 0 & 0 & -1.6818 \\
		\hline
		Euler angle/$^\circ$ & (0,0,135) & (0,0,135) & (7.1099, 4.9624, 134.3901) \\
		\hline
		Scale factor & 1 & 1 & 1.5167 \\
		\hline \hline
	\end{tabular}
	
\end{table}

\section{Conclusion and Discussion}
This paper proposes a differentiable SAR renderer, which can render the 3D model into a 2D SAR image in the forward direction, and inversely obtain the geometric shape and the facet's scattering intensity according to the SAR images in the inverse direction. Different from rendering optical images, differentiable SAR renderer calculates the occlusion through the projection plane, and then formulate the image on the slant-range of the mapping plane. Experiments on targets with ground background, such as building and vehicle, and targets without background, such as ISAR data of satellite and space station, are conducted, which demonstrate the feasibility of inverse graphics for SAR.

When the size of the rendered image is fixed ($128 \times 128$), the time required for DSR rendering is related to the number of facets and the incident angle. As shown in \autoref{table6}, when the facet number increases, the time consumption increases due to the need to interact with more facets; and when the incident angle increases, the number of sampling pixels on the projection plane will also increase, requiring more threads. Compared with the facet number, the increase of the incident angle will have a less impact on time consumption.

\begin{table}[!t]
	\caption{Time consumption of the forward rendering.}
	\label{table6} 
	
	\renewcommand{\arraystretch}{1.5}
	\setlength{\tabcolsep}{1mm}
	\centering
	
	\begin{tabular}{p{1.5cm}<{\centering} p{1.5cm}<{\centering} p{1.5cm}<{\centering} p{1.5cm}<{\centering} p{1.5cm}<{\centering}}
		\hline \hline
		\textbf{Model} & \textbf{Ground} & \textbf{Facet Number} & \textbf{Incident Angle/$^\circ$} & \textbf{Average Time/$s$} \\
		\hline
		\multirow{2} * {Building 1} & \multirow{2} * {$\checkmark$} & \multirow{2} * {5904} & 15 & 0.0933  \\
		\cline{4-5}
		 &  &  & 75 & 0.0958 \\
		\hline
		\multirow{2} * {Building 2} & \multirow{2} * {$\checkmark$} & \multirow{2} * {5904} & 15 & 0.0933 \\
		\cline{4-5}
		 &  &  & 75 & 0.0954 \\
		\hline
		\multirow{2} * {T72 vehicle} & \multirow{2} * {$\checkmark$} & \multirow{2} * {22698} & 15 & 0.2882 \\
		\cline{4-5}
		 &  &  & 75 & 0.2902 \\
		\hline \hline
	\end{tabular}
	
\end{table}

Then observe the time consumption during reconstruction in \autoref{table7}. For the T72 target, when the batch size increases, the average iteration time for each pair of SAR images and silhouettes decreases. When the initial input model of DSR is the same sphere, it takes more time to reconstruct the satellite and Tiangong-1 than to reconstruct T72. This is because the proportion of the two targets in the silhouettes is larger.

\begin{table}[!t]
	\caption{Time consumption of the backward reconstruction.}
	\label{table7} 
	
	\renewcommand{\arraystretch}{1.5}
	\setlength{\tabcolsep}{3.5mm}
	\centering
	
	\begin{tabular}{p{1.5cm}<{\centering} p{1.5cm}<{\centering} p{1.5cm}<{\centering} p{1.5cm}<{\centering}}
		\hline \hline
		\textbf{Target Model} & \textbf{Viewpoint Number} & \textbf{Batch Size} & \textbf{Average Time/$s$} \\
		\hline
		\multirow{3} * {T72 vehicle} & \multirow{3} * {32} & 1 & 0.0696 \\
		\cline{3-4}
		& & 4 & 0.0454 \\
		\cline{3-4}
		& & 8 & 0.0429 \\
		\hline
		Envisat & 32 & 4 & 0.5154\\
		\hline
		Tiangong-1 & 9 & 9 & 0.0818 \\
		\hline \hline
	\end{tabular}
	
\end{table}

\appendices




\ifCLASSOPTIONcaptionsoff
  \newpage
\fi


\begin{thebibliography}{1}
\bibitem{IEEEhowto:Xu_GRSL}
F. Xu, Y. Jin and A. Moreira, "A Preliminary Study on SAR Advanced Information Retrieval and Scene Reconstruction," in IEEE Geoscience and Remote Sensing Letters, vol. 13, no. 10, pp. 1443-1447, Oct. 2016, doi: 10.1109/LGRS.2016.2590878.

\bibitem{IEEEhowto: MicrowaveVision}
C. Ding, X. Qiu, F. Xu, X. Liang, Z. Jiao and F. Zhang, "Synthetic Aperture Radar Three-dimensional Imaging——From TomoSAR and Array InSAR to Microwave Vision", Journal of Radars, vol. 8, no. 6, pp. 693-709, 2019.

\bibitem{IEEEhowto:MPA}
F. Xu and Y. Jin, "Imaging Simulation of Polarimetric SAR for a Comprehensive Terrain Scene Using the Mapping and Projection Algorithm," in IEEE Transactions on Geoscience and Remote Sensing, vol. 44, no. 11, pp. 3219-3234, Nov. 2006, doi: 10.1109/TGRS.2006.879544.

\bibitem{IEEEhowto:BART}
F. Xu and Y. Jin, "Bidirectional Analytic Ray Tracing for Fast Computation of Composite Scattering From Electric-Large Target Over a Randomly Rough Surface," in IEEE Transactions on Antennas and Propagation, vol. 57, no. 5, pp. 1495-1505, May 2009, doi: 10.1109/TAP.2009.2016691.

\bibitem{IEEEhowto:InverseGraphics}
T.D. Kulkarni, V.K. Mansinghka, P. Kohli and J.B. Tenenbaum, "Inverse graphics with probabilistic cad models", arXiv preprint arXiv:1407.1339, 2014.

\bibitem{IEEEhowto:InSARReview}
G. Xu, Y. Gao, J. Li and M. Xing, "InSAR Phase Denoising: A Review of Current Technologies and Future Directions," in IEEE Geoscience and Remote Sensing Magazine, vol. 8, no. 2, pp. 64-82, June 2020, doi: 10.1109/MGRS.2019.2955120.

\bibitem{IEEEhowto:Zhu1}
X. X. Zhu and R. Bamler, "Very High Resolution Spaceborne SAR Tomography in Urban Environment," in IEEE Transactions on Geoscience and Remote Sensing, vol. 48, no. 12, pp. 4296-4308, Dec. 2010, doi: 10.1109/TGRS.2010.2050487.

\bibitem{IEEEhowto:Zhu2}
X. X. Zhu, S. Montazeri, C. Gisinger, R. F. Hanssen and R. Bamler, "Geodetic SAR Tomography," in IEEE Transactions on Geoscience and Remote Sensing, vol. 54, no. 1, pp. 18-35, Jan. 2016, doi: 10.1109/TGRS.2015.2448686.

\bibitem{IEEEhowto:SurveyDGL}
Y. Xiao, Y. Lai, F. Zhang, C. Li and L. Gao, "A survey on deep geometry learning: From a representation perspective", Computational Visual Media, vol. 6, no. 2, pp. 113-133, 2020, doi: 10.1007/s41095-020-0174-8.

\bibitem{IEEEhowto:Pixel2Mesh}
N. Wang, Y. Zhang, Z. Li, Y. Fu, W. Liu and Y. Jiang, "Pixel2mesh: Generating 3d mesh models from single rgb images", In Proceedings of the European Conference on Computer Vision (ECCV), pp. 52-67, 2018, doi: 10.1007/978-3-030-01252-6$\_$4.

\bibitem{IEEEhowto:Pixel2MeshPlus}
C. Wen, Y. Zhang, Z. Li and Y. Fu, "Pixel2mesh++: Multi-view 3d mesh generation via deformation", In Proceedings of the IEEE/CVF International Conference on Computer Vision, pp. 1042-1051, 2019, doi: 10.1109/iccv.2019.00113.

\bibitem{IEEEhowto:Topologies}
J. Tang, X. Han, J. Pan, K. Jia and X. Tong, "A skeleton-bridged deep learning approach for generating meshes of complex topologies from single rgb images", In Proceedings of the IEEE/CVF Conference on Computer Vision and Pattern Recognition, pp. 4541-4550, 2019, doi: 10.1109/cvpr.2019.00467. 

\bibitem{IEEEhowto:PointCloud}
L. Peng, X. Qiu, C. Ding and W. Tie, "Generating 3d Point Clouds from a Single SAR Image Using 3D Reconstruction Network," IGARSS 2019 - 2019 IEEE International Geoscience and Remote Sensing Symposium, 2019, pp. 3685-3688, doi: 10.1109/IGARSS.2019.8900449.

\bibitem{IEEEhowto:SingleTarget}
S. Wang, J. Guo, Y. Zhang, Y. Hu, C. Ding and Y. Wu, "Single Target SAR 3D Reconstruction Based on Deep Learning", Sensors, vol. 21, no. 3, p. 964, 2021, doi: 10.3390/s21030964.

\bibitem{IEEEhowto:3dBuilding}
J. Chen, L. Peng, X. Qiu, C. Ding and Y. Wu, "A 3D building reconstruction method for SAR images based on deep neural network", SCIENTIA SINICA Informationis, vol. 49, no. 12, pp. 1606-1625, 2019, doi: 10.1360/ssi-2019-0100.

\bibitem{IEEEhowto:Opendr}
M. M. Loper and M. J. Black, "Opendr: An approximate differentiable renderer", In European Conference on Computer Vision, pp. 154–169, 2014, doi: 10.1007/978-3-319-10584-0$\_$11.

\bibitem{IEEEhowto:Neural3D}
H. Kato, Y. Ushiku, and T. Harada, "Neural 3d mesh renderer", In Proceedings of the IEEE Conference on Computer Vision and Pattern Recognition, pp. 3907–3916, 2018, doi: 10.1109/cvpr.2018.00411.

\bibitem{IEEEhowto:SoftRas}
S. Liu, T. Li, W. Chen and H. Li, "Soft rasterizer: A differentiable renderer for image-based 3d reasoning", In Proceedings of the IEEE/CVF International Conference on Computer Vision, pp. 7708-7717, 2019, doi: 10.1109/iccv.2019.00780.

\bibitem{IEEEhowto:MSTAR}
Moving and Stationary Target Acquisition and Recognition (MSTAR) Public Release Data. Available online:
https://www.sdms.afrl.af.mil/datasets/matar/

\bibitem{IEEEhowto:flab}
F. Lab. Forscher des fraunhofer fhr begleiten wiedereintritt der
chinesischen raumstation tiangong-1. [Online]. Available: 
https://www.fhr.fraunhofer.de/tiangong-bilder


\end{thebibliography}
\end{document}